\documentclass[10pt,journal,compsoc]{IEEEtran}

\usepackage{ifpdf}
\usepackage{amsmath}

\ifCLASSOPTIONcompsoc
  \usepackage[nocompress]{cite}
\else
  \usepackage{cite}
\fi
\ifCLASSINFOpdf
\else
\fi
\usepackage{amsmath}

\usepackage{mdwmath}
\usepackage{mdwtab}

\usepackage{stfloats}
\usepackage{url}
\usepackage{amssymb}
\usepackage{multirow}
\usepackage[ruled,vlined,linesnumbered]{algorithm2e}
\usepackage{bm}

\usepackage{xcolor}

\usepackage[colorlinks,linkcolor=red,anchorcolor=magenta,citecolor=red]{hyperref}

\usepackage{csquotes}
\usepackage{adjustbox}
\usepackage{booktabs}
\usepackage{mathtools}
\makeatother
\makeatletter

\usepackage{array}
\usepackage[caption=false,font=normalsize,labelfont=sf,textfont=sf]{subfig}
\usepackage{amsfonts}
\usepackage{textcomp}
\usepackage{stfloats}
\usepackage{url}
\usepackage{verbatim}
\usepackage{graphicx}
\usepackage{cite}
\usepackage{booktabs}    
\usepackage{caption}    
\usepackage{multirow}
\usepackage{graphicx} 
\usepackage{subcaption} 
\usepackage{caption}   
\usepackage{amssymb, mathtools}

\usepackage{amsthm}
\newtheorem{definition}{Definition}
\newtheorem{lemma}{Lemma}
\newtheorem{theorem}{Theorem}
\usepackage{enumitem}
\usepackage{hyperref}
\usepackage{lipsum}
\usepackage{indentfirst}
\usepackage{graphicx}

\begin{document}
%
\title{Leveraging Extragradient for Effective Sharpness-Aware Minimization in Deep Learning}
%
%
%
%

\author{Yao~Fu, Chunxia~Zhang, Junmin~Liu, Yihang~Jin, Haishan~Ye, and~Yuanao~Yang%
\IEEEcompsocitemizethanks{%
\IEEEcompsocthanksitem Yao Fu, Junmin Liu, and Chunxia Zhang are with the School of Mathematics and Statistics, Xi’an Jiaotong University, Xi’an, 710049, China (Y. Fu and J. Liu also with SGIT AI Lab, State Grid Corporation of China, Xi’an, 710054, China). 
E-mail: 
fyao56@stu.xjtu.edu.cn., \{junminliu, cxzhang\}@mail.xjtu.edu.cn
\IEEEcompsocthanksitem Haishan Ye is with the School of Management, Xi’an Jiaotong University, Xi’an, 710049, China; SGIT AI Lab, State Grid Corporation of China, Xi’an, 710054, China. E-mail: hsye\_cs@outlook.com.%
\IEEEcompsocthanksitem Yihang Jin is with the School of Software, Xi’an Jiaotong University, Xi’an, 710049, China. E-mail: yihangjin@stu.xjtu.edu.cn.%
\IEEEcompsocthanksitem Yuanao Yang is with the State Key Laboratory of Multiphase Flow in Power Engineering, Xi’an Jiaotong University, Xi’an, 710049, China. E-mail: yya0407@stu.xjtu.edu.cn.%
}         } 
\IEEEtitleabstractindextext{%
\begin{abstract}
Generalization remains a pivotal challenge in deep learning, where traditional optimizers like Stochastic Gradient Descent (SGD) often converge to sharp minima, leading to overfitting and reduced performance on unseen data. Building on Sharpness-Aware Minimization (SAM), for seeking flat minima associated with improved generalization, we propose the Extragradient-Inspired Sharpness-Aware Minimization (EISAM), a novel optimizer that enhances generalization via the extragradient technique. EISAM uses a two-step update process: a prediction step investigating the geometry of the loss landscape and a perturbation step that refines updates with a base optimizer. This approach achieves better generalization performance than SAM. Crucially, EISAM reduces sensitivity to the perturbation radius, enhancing robustness, and simplifying the tuning across diverse settings. Extensive experiments on benchmark datasets demonstrate that EISAM consistently outperforms SGD, Adaptive Moment Estimation (Adam), and SAM in test accuracy and training efficiency across various architectures. Theoretical analysis further confirms that EISAM tightens the generalization bound by steering parameters toward flatter minima with reduced curvature. Accompanied by a thorough hyperparameter analysis, EISAM offers practical tuning guidance, establishing it as a robust, scalable, and broadly applicable optimization solution that advances both the theory and practice in deep learning.
\end{abstract}

}

\maketitle

\IEEEdisplaynontitleabstractindextext

%
\IEEEpeerreviewmaketitle

\ifCLASSOPTIONcompsoc
\IEEEraisesectionheading{\section{Introduction}\label{sec1}}
\else
\section{Introduction}
\label{sec1}
\fi

%
%
%
%
\IEEEPARstart{O}{ptimizing} deep neural networks to achieve better generalization performance is a key challenge in deep learning. Traditional optimizers, such as SGD and Adam, have achieved significant success in training deep models, but they often converge to sharp minima of the loss function. Although the association between flat minima and generalization is controversial \cite{Dinh2017}, studies have shown that sharp minima often lead to poor generalization on unseen data, while flat minima are associated with stronger generalization performance \cite{Hochreiter1997, Keskar2017}. Building on these insights, SAM \cite{Foret2021} has emerged as an effective approach to seek flat minima by minimizing loss sharpness within a parameter neighborhood. Despite its benefits, SAM requires additional gradient computations at each step, resulting in approximately twice the computational cost compared to base optimizers like SGD and Adam, though this can be mitigated through faster convergence. Moreover, its performance is highly sensitive to the perturbation radius $\rho$, and the sharpness approximation may not accurately capture the loss landscape's curvature, potentially yielding suboptimal generalization.

The extragradient method is an optimization technique designed to solve problems in which the objective function combines a smooth part and a convex part \cite{Nguyen2018}. Unlike traditional gradient descent, which uses one step per iteration, this method takes two steps: a prediction step to explore the problem's structure and a correction step to adjust the direction. This makes it especially useful for tasks with non-smooth elements, such as signal processing or machine learning. It is robust in choosing step sizes, stable in practice, and can guarantee convergence to a critical point under certain conditions. For convex problems, it offers a sublinear convergence rate and can even achieve linear convergence in some cases. Its flexibility also allows it to tackle both convex and nonconvex challenges, making it a powerful and adaptable tool for various applications.

\begin{figure}[h]
    \centering 
    \includegraphics[width=0.5\textwidth, trim=0.0cm 0.0cm 0.0cm 0.0cm, 
    clip,angle=0]{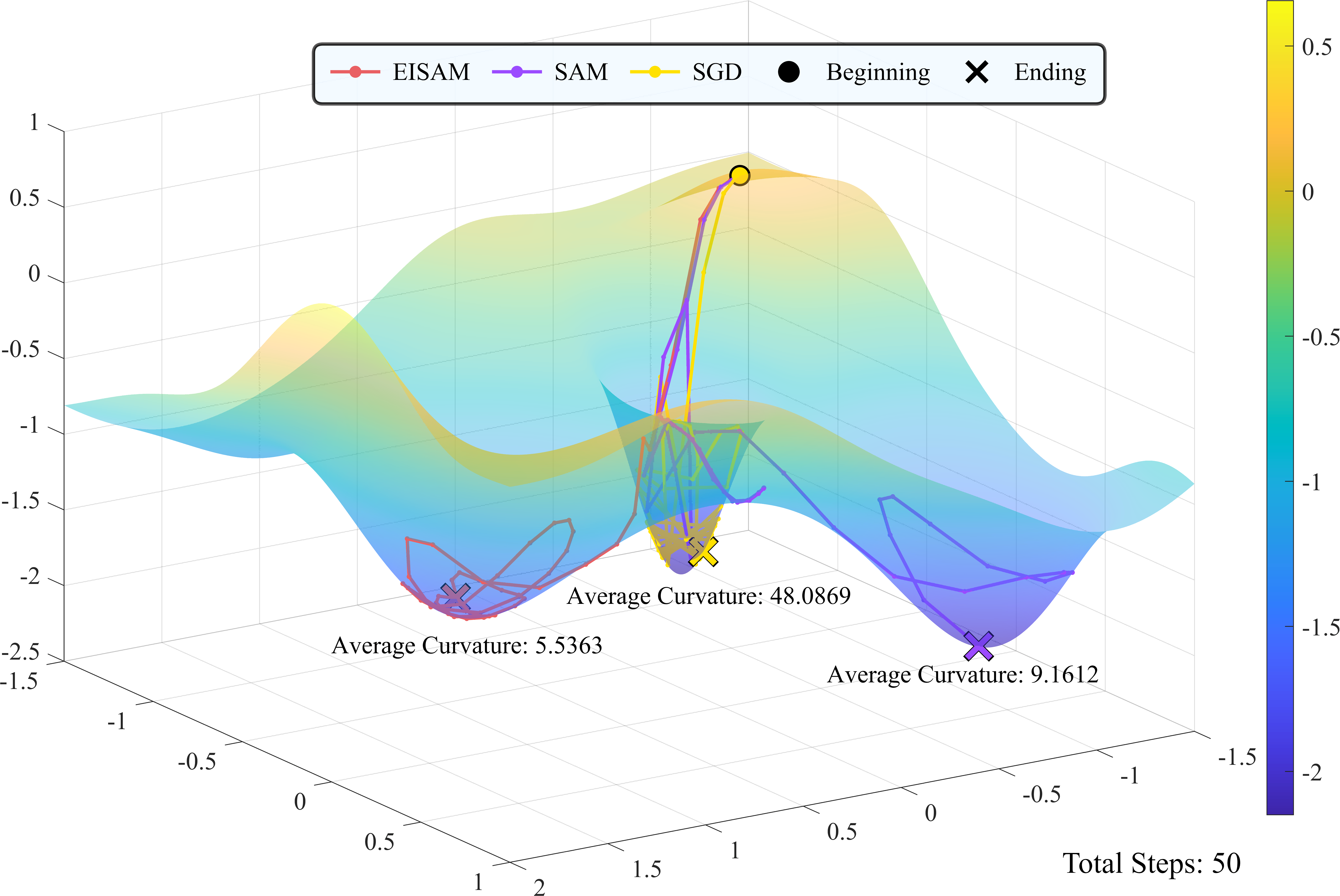}	
        \caption{Visualization of training trajectories in the loss landscape. Lower average curvature indicates flatter minima and better expected generalization. EISAM achieves lower curvature than SAM and SGD.} 
    \label{3d}
\end{figure}

We propose EISAM, a novel optimizer designed to address the limitations of SAM, particularly in terms of generalization and hyperparameter sensitivity. EISAM refines SAM’s perturbation and parameter update mechanisms via a two-step strategy: a prediction step that probes the local geometry of the loss landscape, followed by a correction step performed with a base optimizer. This approach improves generalization to unseen data, reduces dependence on precise tuning of the perturbation radius $\rho$, and guides the optimization trajectory toward flatter minima with enhanced robustness. Consequently, EISAM not only bolsters training stability but also broadens its applicability across a diverse range of machine learning tasks and domains.

To illustrate the motivation behind our approach, consider a conceptual example of the optimization trajectory in a representative loss landscape (see Fig. \ref{3d}). Traditional SGD converges to sharp minima with high curvature, leading to poor generalization. While SAM attempts to mitigate this by seeking flatter regions through worst-case perturbations, it can still result in trajectories that settle in sub-optimal flat areas or exhibit instability due to sensitivity to the perturbation radius $  \rho  $. In contrast, by incorporating an extragradient-inspired prediction step that proactively explores the local geometry before applying the perturbation correction, our proposed EISAM steers the optimizer toward flatter minima with broader basins and lower curvature, effectively serving as an adaptive regularizer along the optimization trajectory. This two-step mechanism intuitively promotes more robust and stable convergence to superior generalization solutions. Further optimizer comparisons from various perspectives are presented in the Appendix \ref{appa}.

Our contributions are as follows:
\begin{itemize}
    \item \textbf{We propose a novel optimizer EISAM:} Inspired by the extragradient method, EISAM improves the computational flow of SAM while enhancing generalization performance and providing robustness to SAM's key hyperparameter $\rho$.
    \item \textbf{Algorithmic and Theoretical Foundations:} A detailed description of EISAM's algorithmic design is presented, accompanied by theoretical analysis that elucidates how the extragradient-inspired two-step update strategy achieves effective sharpness-aware optimization and tightens the generalization bound.    
    \item \textbf{Experimental Validation:} Extensive experiments on CIFAR-10, CIFAR-100 \cite{krizhevsky2009cifar}, ImageNet-1K \cite{deng2009imagenet}, COCO \cite{lin2014coco}, LVIS v1.0 \cite{gupta2019lvis}, ISIC2018 \cite{codella2019isic}, and BOOLQ \cite{clark2019boolq} datasets, demonstrate that EISAM exhibits better performance in terms of generalization.
    \item \textbf{Hyperparameter Analysis:} The selection of key hyperparameters in EISAM and their impact on model performance are discussed, providing practical guidance for real-world applications.
\end{itemize}

Through these improvements, EISAM opens new possibilities for deep learning optimization. Under computational budgets comparable to those of competing methods, EISAM attains better model performance. It offers researchers and practitioners an optimization tool that balances performance and efficiency.

\section{Related Work}
\label{sec2}
The design of optimization algorithms in deep learning is intricately linked to the generalization performance of neural networks. Recent research has emphasized the role of the loss function’s geometric properties in shaping model outcomes \cite{Andriushchenko2022, Dinh2017}. This section reviews key literature on flat minima theory, the evolution of SAM and its variants, theoretical underpinnings of optimization algorithms, network architecture design, data augmentation strategies, and advances in optimization techniques, providing a comprehensive foundation for this study.

\textbf{Flat Minima Theory.} The idea of targeting flat minima in optimization was pioneered by Hochreiter and Schmidhuber \cite{Hochreiter1997}, who argued that regions in the parameter space with flatter loss surfaces lead to better generalization. This theory has gained substantial support over time. For example, Chaudhari et al. \cite{Chaudhari2017} introduced Entropy-SGD, which uses an entropy term to steer gradient descent toward wide valleys, yielding improved test performance in tasks like image classification and natural language processing. Keskar et al. \cite{Keskar2017} linked this debate to training dynamics, noting that large-batch training often pushes models toward sharp minima, correlating with reduced generalization. Additionally, Li et al. \cite{Li2018} examined loss function geometry, suggesting that flat minima may arise from the over-parameterization of deep networks, a feature that could explain the success of optimization methods in high-dimensional spaces.

However, the supremacy of flat minima has been debated. Dinh et al. \cite{Dinh2017} demonstrated that sharp minima can also generalize well under specific conditions, challenging the notion that flatness is always preferable. Notwithstanding these debates, substantial experimental evidence indicates that flatter minima are consistently associated with improved generalization in practical settings \cite{andriushchenko2023modern, schliserman2025flat, lell2023split}. These theoretical insights motivate sharpness-aware optimizers, but many early methods, such as Entropy-SGD, suffer from high computational overhead due to nested iterations, limiting their scalability to large models. Empirical studies in specialized domains, such as medical imaging, confirm that sharpness-aware optimizers like SAM enhance generalization, though variants require domain-specific refinements \cite{hassan2025sharpness}, supporting EISAM's applicability in diverse tasks.

\textbf{Sharpness-Aware Minimization and its Variants.} Building on flat minima theory, Foret et al. \cite{Foret2021} proposed SAM, an algorithm that minimizes loss sharpness by regularizing the maximum loss within a parameter neighborhood. SAM has delivered notable generalization gains on datasets like CIFAR-10 and ImageNet-1K. This success spurred refinements, such as Adaptive Sharpness-Aware Minimization (ASAM) \cite{Kwon2021} which adjusts sharpness dynamically for different tasks and architectures, and Sharpness-Aware Lookahead (SALA) \cite{Tan2024} incorporates SAM into Lookahead's framework, mitigating Lookahead's tendency toward suboptimal generalization while preserving its acceleration benefits. Furthermore, Efficient Sharpness-Aware Minimization (ESAM) \cite{Du2020} improves the computational efficiency while retaining SAM’s generalization benefits. Friendly Sharpness-Aware Minimization (FSAM) \cite{du2024friendly} focuses on efficiency by decomposing the adversarial perturbation into full gradient and stochastic gradient noise components, removing the full gradient (estimated via EMA) to mitigate inconsistencies and enhance generalization without extra computational overhead. Similarly, Surrogate Gap Guided Sharpness-Aware Minimization (GSAM) \cite{zhuang2022surrogate} improves generalized sharpness by introducing a surrogate gap minimization step in an orthogonal direction after gradient descent, which further reduces sharpness while maintaining low loss. These methods achieve better efficiency and broader sharpness awareness compared to vanilla SAM. Recent theoretical advancements provide unified analyses of SAM and its variants, demonstrating improved convergence rates under relaxed noise assumptions \cite{oikonomou2025sharpness}. 

Despite these advancements, SAM and its variants have limitations. SAM requires additional gradient computations, making it computationally intensive for large-scale datasets and models. Moreover, its performance is highly sensitive to the perturbation radius $\rho$, necessitating careful hyperparameter tuning, which can hinder practical deployment across diverse settings. ESAM \cite{Du2020} mitigates efficiency issues to some extent but still inherits sensitivity to hyperparameters and may not fully capture the loss landscape's curvature for optimal generalization. These shortcomings highlight the need for a more robust and efficient optimizer. Hence, we propose EISAM to address these issues by leveraging an extragradient-inspired two-step update strategy, reducing the burden of hyperparameter tuning and hyperparameter sensitivity while achieving better generalization.

\textbf{Theoretical Underpinnings of Optimization Algorithms.} The theoretical basis for SAM and EISAM builds upon foundational optimization research. Nesterov \cite{Nesterov1983} introduced accelerated gradient descent, incorporating momentum for faster convergence, which set the stage for contemporary methods. Bolte et al. \cite{Bolte2017} analyzed the complexity of first-order descent methods for convex functions, providing a framework to assess the efficiency of algorithms. Generalization analysis has also leaned on the PAC-Bayes framework \cite{McAllester1999}. Jiang et al. \cite{Jiang2019} compared generalization metrics like sharpness and parameter norms, offering empirical support.
These theoretical tools enable rigorous analysis of convergence and generalization bounds, but applying them to sharpness-aware methods often reveals gaps, such as unaddressed curvature effects in non-convex landscapes. In contrast, EISAM extends this by theoretically tightening generalization bounds through flatter minima with reduced curvature.

\textbf{Network Architecture Design.} Optimization efficacy often depends on network architecture. ResNet introduced residual connections to mitigate vanishing gradients \cite{He2016}, enhancing training efficiency and creating a solid platform for testing optimizers. PyramidNet \cite{Han2017} incrementally widens network layers to increase capacity, while WideResNet \cite{Zagoruyko2016} boosts performance through broader layers. These architectural innovations provide varied environments for evaluating optimization strategies, but they can exacerbate computational challenges in sharpness-aware methods.

\textbf{Data Augmentation and Regularization.} Data augmentation and regularization are vital for generalization. AutoAugment automates augmentation policy learning \cite{Cubuk2019}, improving classification performance. CutMix \cite{Yun2019} blends image regions and labels to enhance robustness. Muller et al. \cite{Muller2019} demonstrated that moderate label smoothing serves as an effective regularizer that reduces overfitting and enhances generalization. Similarly, Mixup interpolates samples and labels, complementing sharpness-aware optimization \cite{Zhang2018}.
These techniques synergize with sharpness-aware approaches but often require additional tuning to avoid over-regularization. 

\textbf{Advances in Optimization Techniques.} Optimization innovations have revitalized deep learning. Kingma and Ba introduced Adam \cite{Kingma2015}, which uses adaptive learning rates and is widely used in training deep networks. Loshchilov and Hutter \cite{Loshchilov2017} employed SGDR-learning rate restarts to boost convergence and generalization. Lookahead \cite{Zhang2019} optimizer uses fast and slow parameter updates for stability, inspiring multi-step approaches like EISAM. RAdam refines Adam’s variance estimation for greater robustness \cite{Scabini2023}.

While these methods improve convergence, they may converge to sharp minima, limiting generalization. EISAM draws inspiration from multi-step updates to refine sharpness-aware optimization, balancing efficiency and performance. Prior work on optimization algorithms, flat minima theory, and generalization analysis has laid a robust foundation for SAM and EISAM. This paper builds on these insights, investigating EISAM’s convergence properties and its impact on neural network generalization, contributing new perspectives to the theory and application of optimization.

\section{Methodology}
\label{sec3}

\begin{figure}[t]
    \centering 
    \includegraphics[width=0.5\textwidth, trim=0.3cm 0cm 0.1cm 0.1cm, 
    clip,angle=0]{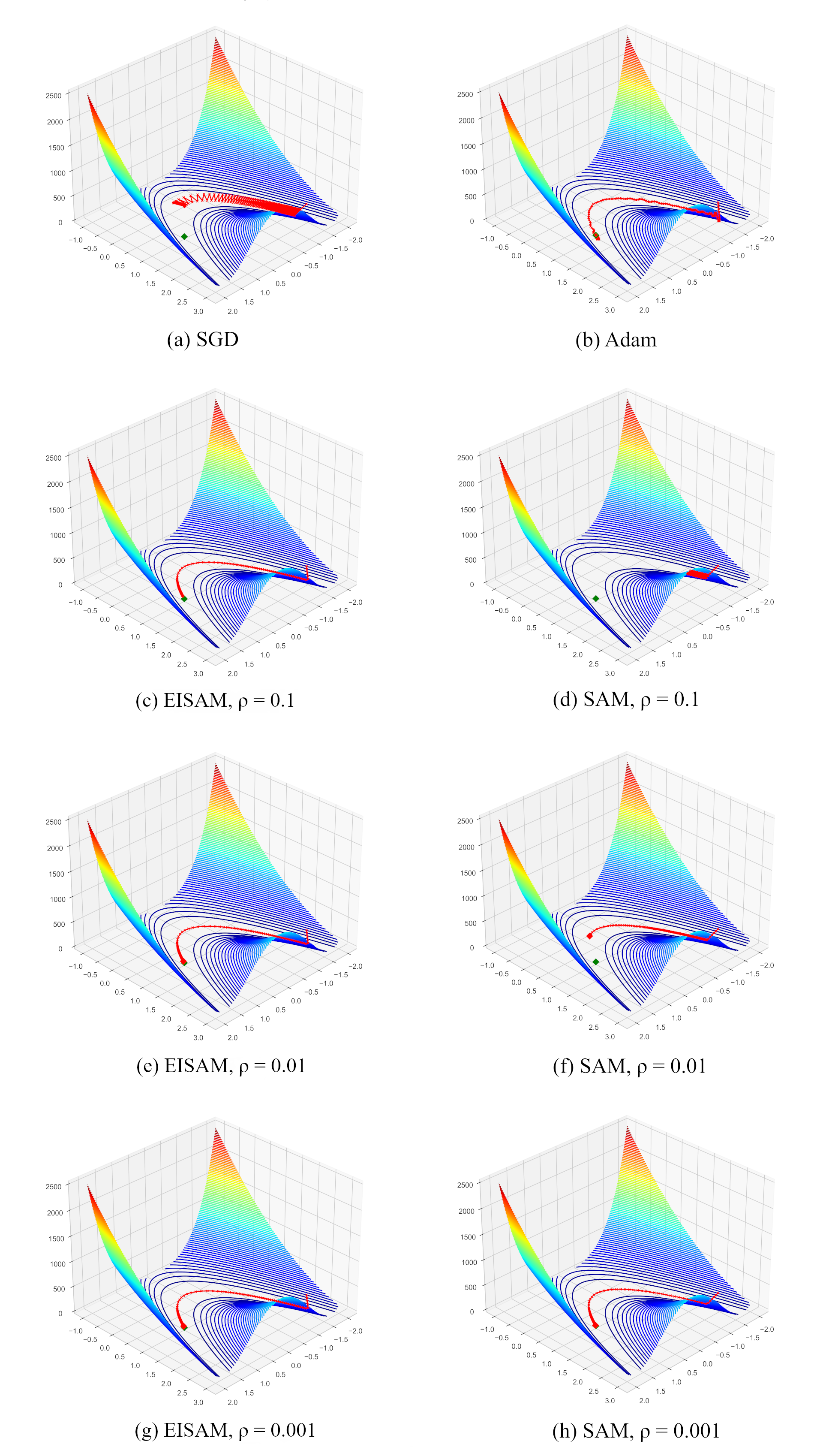}	
    \caption{Optimization trajectories of various optimizers on the Rosenbrock function, visualized in 3D contour plots. The color map represents the loss surface, with red lines indicating the optimization paths from the initial point (-2.0, 2.0) to the global minimum (1.0, 1.0) over 200 iterations. Subfigures show: (a) SGD, (b) Adam, (c) EISAM with $\rho$=0.1, (d) SAM with $\rho$=0.1, (e) EISAM with $\rho$=0.01, (f) SAM with $\rho$=0.01, (g) EISAM with $\rho$=0.001, and (h) SAM with $\rho$=0.001.} 
    \label{F1}%
\end{figure}

To illustrate how various optimizers navigate the loss landscape and to highlight EISAM's robustness to the perturbation radius $  \rho  $, Fig. \ref{F1} shows 3D visualizations of optimization trajectories on the Rosenbrock function \cite{Rosenbrock1960AnAM}, comparing SGD, Adam, SAM, and EISAM. It can be observed in Fig. \ref{F1} that SGD exhibits a broad descent path, reflecting its reliance on pure local gradient information without adaptive mechanisms. The trajectory gradually spirals toward minimum, indicating stable but slow convergence. This behavior often leads SGD to broad, flat regions, which can promote generalization in certain scenarios, yet it risks settling in suboptimal areas due to limited precision in complex, curved landscapes. In contrast, Adam demonstrates a smoother and more direct trajectory with reduced oscillations. By leveraging adaptive step sizes through moving averages of gradients and squared gradients, Adam achieves faster and more stable convergence compared to SGD. The resulted minimum strikes a balance between flatness and depth. However, this accelerated descent frequently comes at the cost of slightly compromised generalization performance.

SAM employs an adversarial perturbation to minimize loss sharpness, guiding the optimization trajectory toward flatter regions. However, its performance is sensitive to the perturbation radius $\rho$: smaller values enable effective exploration, while large values often result in dispersed trajectories. Our proposed EISAM leverages an extragradient-inspired two-step prediction-correction mechanism to refine this process. EISAM produces tighter and more concentrated trajectories across the same range of $\rho$ values. This reduces sensitivity to $  \rho  $, yielding lower hyperparameter tuning burden and superior generalization, making EISAM a more robust sharpness-aware optimizer for challenging landscapes such as the Rosenbrock function.

\subsection{Sharpness-Aware Minimization}
\label{3.1}

Empirical studies have consistently shown that solutions residing in flat regions of the loss landscape tend to generalize better to unseen data than those in sharp minima. These flat regions are characterized by minimal variation in the loss function under small perturbations, making the corresponding solutions less sensitive to parameter changes and thus less prone to overfitting. SAM addresses this by explicitly incorporating a sharpness measure into the optimization process. It quantifies sharpness by solving a maximization problem that seeks perturbations within a neighborhood of radius $\rho$ to maximize the local loss elevation. Throughout this paper, we denote a neural network $f$ with parameters $\mathbf{w}$ as $f_{\mathbf{w}}$, where $\mathbf{w}=(w_1,w_2,\cdots,w_N)^T$ and $N$ is the total number of weight units. For a given parameter vector $\mathbf{w}$ and an empirical risk $F_S(\mathbf{w})$ on training set $S$, SAM aims to find:
\begin{align}
\boldsymbol{\epsilon}^* = \arg\max_{\|\boldsymbol{\epsilon}\|_2 \leq \rho} F_S(\mathbf{w} + \boldsymbol{\epsilon}).
\end{align}
In this formulation, $  \boldsymbol{\epsilon}  $ is a perturbation vector satisfying $  \|\boldsymbol{\epsilon}\|_2 \leq \rho  $, and $  \boldsymbol{\epsilon}^*  $ denotes the optimal (worst-case) perturbation that maximizes the empirical risk $  F_S  $ over the $  \ell_2  $-ball centered at $  \mathbf{w}  $. However, solving this exactly would be computationally intensive, as it requires evaluating the loss multiple times. To make it practical, SAM approximates $\boldsymbol{\epsilon}^*$ using a first-order Taylor expansion of $F_S(\mathbf{w} + \boldsymbol{\epsilon})$, that is,
\begin{align*}
\max_{\|\boldsymbol{\epsilon}\|_2\leq \rho}F_S(\mathbf{w} + \boldsymbol{\epsilon}) \approx \max_{\|\boldsymbol{\epsilon}\|_2\leq \rho}F_S(\mathbf{w}) + \boldsymbol{\epsilon}^\top \nabla F_S(\mathbf{w}),
\end{align*}
leading to the approximate solution
\begin{align*}
\boldsymbol{\epsilon}^* \approx \rho \frac{\nabla F_S(\mathbf{w})}{\|\nabla F_S(\mathbf{w})\|_2}.
\end{align*}
This perturbation is in the direction of the gradient ascent, scaled to lie on the boundary of the $\rho-ball$. With this approximation, SAM computes the gradient at the perturbed point $\mathbf{w} + \boldsymbol{\epsilon}^*$ and uses it to update the original parameters $\mathbf{w}$:
\begin{align}
\mathbf{w} \leftarrow \mathbf{w} - \eta \nabla F_S(\mathbf{w} + \boldsymbol{\epsilon}^*),
\end{align}
where $\eta$ is the learning rate. This update biases the optimization trajectory towards regions where the loss remains low even under perturbations, promoting flatter minima.

The algorithmic implementation of SAM is straightforward and can be integrated with base optimizers like SGD or Adam. In practice, SAM alternates between computing the perturbed gradient and performing the update, often requiring an additional forward and backward pass per iteration. Despite this, its simplicity allows seamless incorporation into existing training pipelines. Extensive experiments on various image tasks, such as image classification, have demonstrated that SAM produces models with superior generalizability \cite{Foret2021}.

\subsection{Overview of the Extragradient Method}
\label{3.2}

The Extragradient method, originally proposed by Korpelevich \cite{Korpelevich1976}, is a classical optimization technique initially developed for solving variational inequality problems. In the context of optimization, it has gained attention for its ability to handle composite objective functions, particularly those involving a smooth function combined with a nonsmooth convex term, as in the form:
\begin{align}
    \min_{\mathbf x \in \mathbb{R}^n} F({\mathbf{x}}) := f({\mathbf{x}}) + g({\mathbf{x}}),
\end{align}
where $ f $ is a differentiable function with an $ L $-Lipschitz continuous gradient, and $ g $ is a proper, lower semicontinuous, convex function. Unlike standard gradient-based methods that rely on a single gradient evaluation per iteration, the extragradient method employs a two-step update strategy. This special strategy enhances its robustness by incorporating curvature information about the loss landscape.

The core idea of the extragradient method is to perform two proximal gradient steps at each iteration. Given an initial point $ \mathbf x_k \in \mathbb{R}^n $, the method generates an intermediate point $ \mathbf y_k $ and updates the solution $ \mathbf x_{k+1} $ as follows:
\begin{align*}
    \begin{cases}
    \mathbf y_k := \operatorname{prox}_{s_k g} \left( \mathbf x_k - s_k \nabla f(\mathbf x_k) \right), \\
    \mathbf x_{k+1} := \operatorname{prox}_{\alpha_k g} \left( \mathbf x_k - \alpha_k \nabla f(\mathbf y_k) \right),
    \end{cases}
\end{align*}
where $ \operatorname{prox}_{t g} $ is the proximal operator associated with the function $ g $ and the parameter $t$, defined as:
\begin{align*}
\operatorname{prox}_{t g}(\mathbf x) := \operatorname{argmin}_{\mathbf y \in \mathbb{R}^n} \left\{ g(\mathbf y) + \frac{1}{2 t} \| \mathbf y - \mathbf x \|^2 \right\},
\end{align*}
and $ s_k, \alpha_k > 0 $ are step sizes satisfying specific conditions, such as $ s_k \leq \alpha_k $ and $ s_k < \frac{1}{L} $, to ensure convergence \cite{Luo1993}, where $L > 0$ denotes the Lipschitz constant of the gradient. The first step computes $ \mathbf y_k $ by taking a proximal gradient step from $ \mathbf x_k $, effectively probing the geometry of the problem. The second step updates $ \mathbf x_{k+1} $ using the gradient evaluated at $ \mathbf y_k $, which provides a more informative descent direction compared to the standard gradient descent.

The extragradient method’s dual-step approach offers several advantages. First, it leverages the additional gradient evaluation at $ \mathbf y_k $ to better capture the curvature of the loss landscape, potentially allowing for larger step sizes and faster convergence. Second, under the Kurdyka-Łojasiewicz (KL) assumption, the method guarantees convergence to a critical point in nonconvex settings and exhibits a finite-length sequence property, meaning the total distance traveled by the iterates is bounded \cite{Attouch2009, Attouch2013, Bolte2014}. In convex settings, it achieves a sublinear convergence rate of $ O(1/T) $, where $ T $ is the number of iterations, akin to classical first-order methods \cite{Beck2009}.

Extragradient methods have been extended to deep learning contexts, such as EGPO for Nash equilibria in preference optimization, offering last-iterate convergence guarantees \cite{zhou2025extragradient}, which inspires EISAM's two-step updates for flatter minima. Despite its strong theoretical foundations, including effective exploration of the loss geometry via a two-step prediction-correction process and guaranteed sublinear (or linear) convergence in convex composite optimization, the extragradient method needs further adaptation for deep learning applications. These advantages motivate us to propose the EISAM optimizer, which builds upon the extragradient's two-step framework and integrates sharpness-aware minimization techniques, improving the model's generalization ability by guiding parameters to converge towards flatter minima.

\subsection{Extragradient Inspired Sharpness-Aware Minimization}
\label{3.3}

EISAM extends the SAM framework by incorporating an extragradient-inspired two-step mechanism: a prediction step that probes the loss landscape ahead of the current parameters, followed by a refined perturbation step. This iterative exploration enables more informative updates, effectively guiding the optimization toward flatter regions with enhanced robustness. Formally, given a parameter vector $\mathbf{w}_t$ at the $t$-th iteration and the empirical risk $F_S(\mathbf{w}_t)$ over a mini-batch, EISAM proceeds as follows. First, it computes an intermediate point $\mathbf{y}_t$ via a prediction step:
\begin{align}
\mathbf{y}_t = \mathbf{w}_t - s \nabla F_S(\mathbf{w}_t),
\end{align}
where $s > 0$ is the prediction step size. This step acts as a lookahead, exploring the negative gradient direction to identify a potentially advantageous position. Next, EISAM performs a sharpness-aware perturbation at the intermediate prediction vector $  \mathbf{y}_t  $, which computes the gradient at $  \mathbf{y}_t  $, yielding:
\begin{align*}
\boldsymbol{\epsilon}_t = \rho \frac{\nabla F_S(\mathbf{y}_t)}{\|\nabla F_S(\mathbf{y}_t)\|_2}.
\end{align*}
This exact two-step mechanism, inspired by the classical extragradient method, effectively probes the local curvature and supports our theoretical claims of steering toward flatter minima with reduced sensitivity to $  \rho  $. 
While an exact approach would require computing $\nabla F_S(\mathbf{y}_t)$, which would add one extra gradient computation, EISAM approximates it using $\nabla F_S(\mathbf{w}_t)$ for efficiency, actually simulates three gradient computations. This simulation, besides introducing diverse new information during the training process to improve generalization, also avoids mechanical paths like those produced by SAM when encountering complex problems (for example, repeatedly oscillating when trapped in saddle points) yielding:
\begin{align*}
\boldsymbol{\epsilon}_t \approx \rho \frac{\nabla F_S(\mathbf{w}_t)}{\|\nabla F_S(\mathbf{w}_t)\|_2},
\end{align*}
where $\rho > 0$ is the neighborhood radius. The perturbed point is then $\mathbf{y}_t + \boldsymbol{\epsilon}_t$, and the update uses the gradient evaluated there:
\begin{align}
\mathbf{w}_{t+1} = \mathbf{y}_t - \eta \nabla F_S(\mathbf{y}_t + \boldsymbol{\epsilon}_t),
\end{align}
with $\eta > 0$ denoting the learning rate. This process biases the optimization toward flatter minima by incorporating landscape information from the predicted point $\mathbf{y}_t$.

\begin{algorithm}[!t]
\SetAlgoLined
\SetKwInOut{Require}{Require}
\SetKwInOut{Ensure}{Ensure}
\KwIn{Network $f_{\mathbf{w}}$ with parameters $\mathbf{w} = (w_1, w_2, \ldots, w_N)^T$, Training set $S$, Batch size $b$, Learning rate $\eta > 0$, Perturbation radius $\rho > 0$, Prediction step size $s > 0$, Number of iterations $T$, Base optimizer $\textit{opt}$ (e.g., SGD, Adam)}
\KwOut{Optimized parameters $\hat{\mathbf{w}}$}
\For{$t=1$ to $T$}{
    Sample a mini-batch $B \subset S$ with size $b$\;
    Compute the gradient $g = \nabla_{\mathbf{w}} F_B(\mathbf{w})$\;
    Compute the intermediate point $\mathbf{y} = \mathbf{w} - s \cdot g$, update parameters $\mathbf{w}\leftarrow \mathbf{y}$\;
    Compute the perturbation $\epsilon = \rho \frac{g}{\|g\|_2}$\;
    Compute the sharpness-aware gradient $g_{\mathrm{sam}} = \nabla_{\mathbf{w}} F_{B}(\mathbf{w})$\;
    Update parameters $\mathbf{w} \leftarrow \textit{opt}.update(\mathbf{w}, g_{\mathrm{sam}}, \eta)$\;
}
\caption{EISAM}
\label{alg:eisam}
\end{algorithm}

The algorithmic implementation of EISAM is compatible with base optimizers like SGD or Adam and can be readily integrated into standard training workflows. In practice, EISAM requires computing gradients at two points per iteration, akin to SAM, but the predictive exploration may facilitate escaping sharp regions more effectively. 

\begin{figure}[htbp]
    \centering 
    \includegraphics[width=0.5\textwidth, trim=25cm 16cm 30cm 12cm, 
    clip,angle=0]{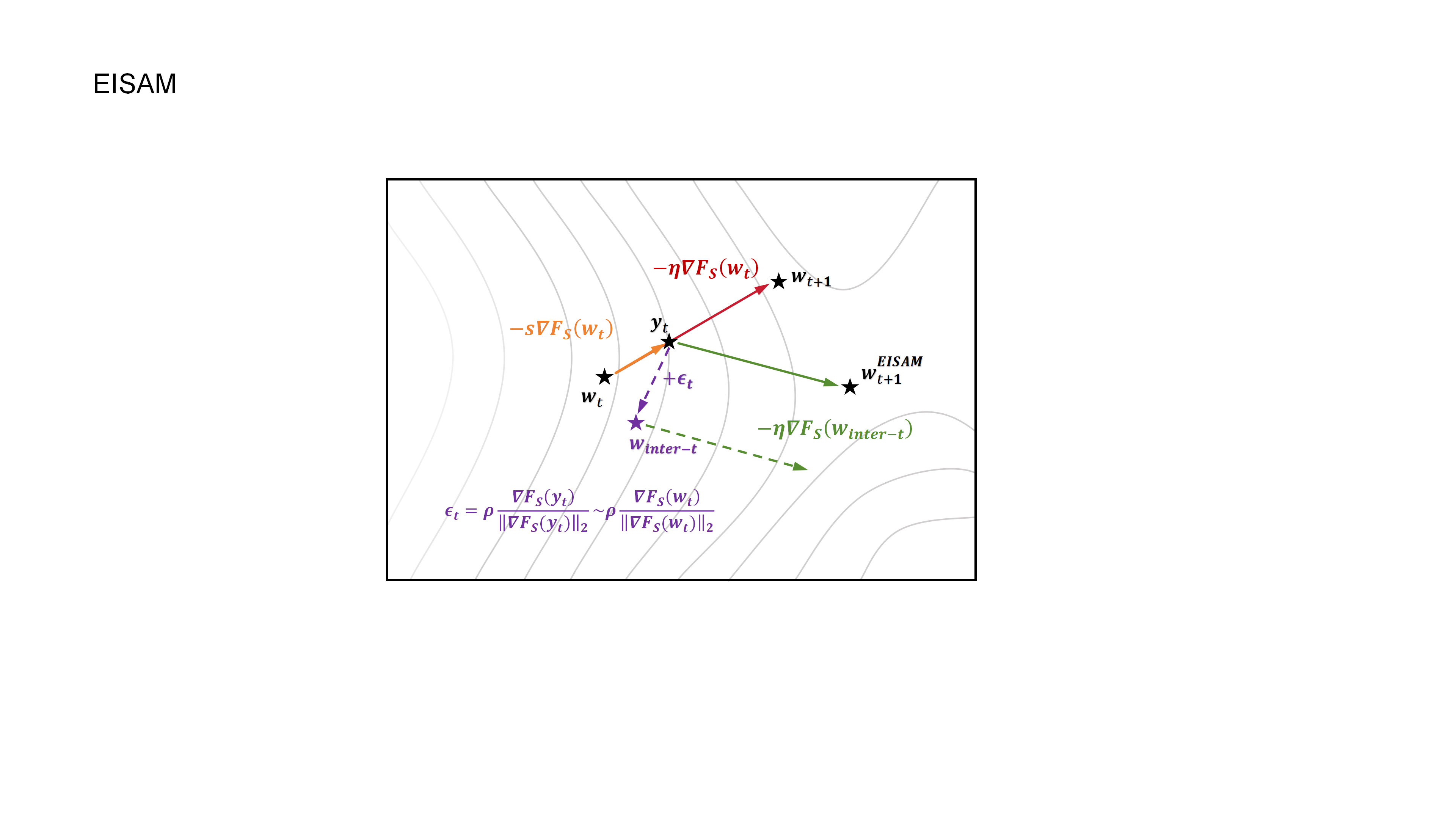}	
    \caption{Schematic of the EISAM parameter update, $\mathbf{w}_{inter-t}$: intermediate adversarial (perturbed) point from SAM, used only for gradient evaluation and not actually updated.} 
    \label{F2}
\end{figure}

Although SAM and EISAM both require two full gradient evaluations per optimization step, EISAM demonstrates superior training efficiency in practice. Its extragradient-inspired prediction step enables advance exploration of the loss landscape's geometry, while SGD's strength lies in its implicit regularization. Put in another way, the prediction step introduces effective noise that biases the optimization path toward flat regions. Additionally, EISAM exhibits faster convergence, greater robustness to $  \rho  $, and reduced hyperparameter tuning needs, collectively shortening total training time while preserving or improving generalization over SAM. The practical implementation of EISAM is summarized in Algorithm \ref{alg:eisam}, and Fig. \ref{F2} schematically illustrates a single EISAM parameter update. Compared to SAM, which directly perturbs based on the gradient at $\mathbf{w}_t$, EISAM's prediction step introduces a lookahead mechanism that leverages local geometry for potentially more robust perturbations. This may enable EISAM to converge to flatter minima with improved generalization. 

\subsection{Hessian Spectra}
\label{3.4}

\begin{figure*}[htbp]
    \centering 
    \includegraphics[width=1\textwidth, trim=0cm 0cm 0cm 0cm, 
    clip,angle=0]{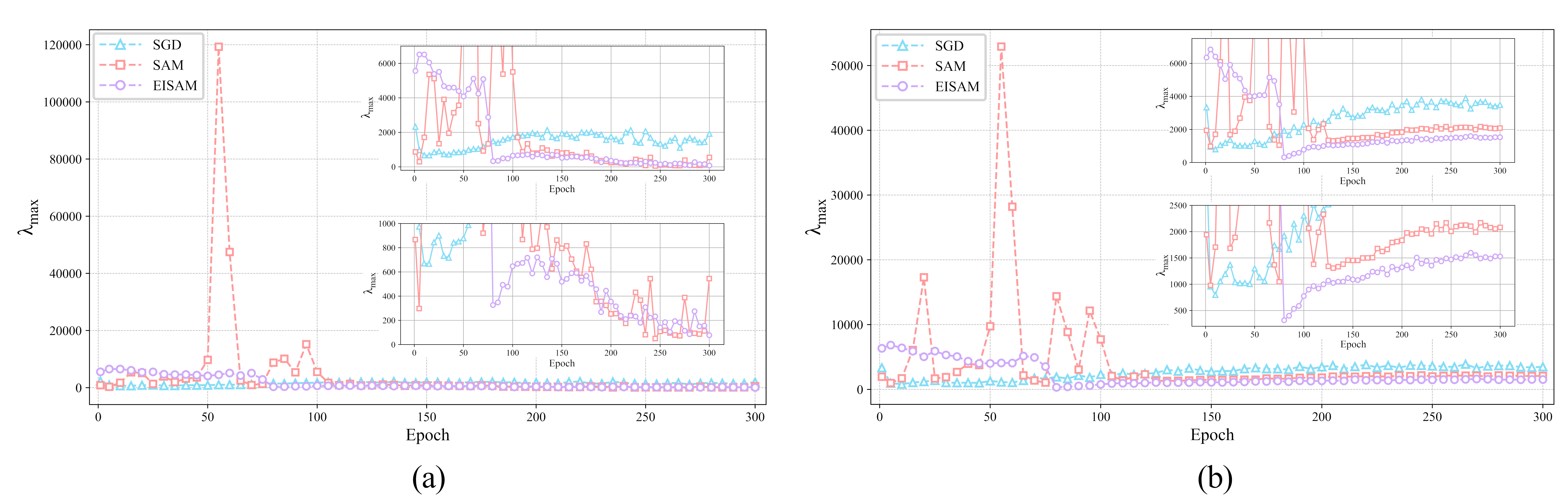}	
    \caption{Trends of the average maximum eigenvalue ($\lambda_{\max}$) across epochs for different optimizers using a ResNet50 model on the CIFAR-100. The blue, red and purple lines correspond to SGD, SAM and EISAM, respectively. This figure is based on the average results from three experiments with different random seeds: (a) Average $\lambda_{\max}$ trend on the training set, (b) Average $\lambda_{\max}$ trend on the test set.} 
    \label{F3}%
\end{figure*}

Motivated by the connection between the geometry of the loss landscape and generalization \cite{Keskar2017}, we conducted experiments to analyze the curvature properties of minima found by different optimizers. Specifically, we trained a ResNet50 model on the CIFAR-100 dataset using three optimization methods: SGD, SAM, and EISAM. To assess the curvature of the loss landscape, we computed the average maximum eigenvalue $\lambda_{\max}$ of the Hessian matrix at various epochs during training. To avoid the confounding effects of batch normalization on Hessian interpretation, we removed batch normalization layers from the model. Given the high dimensionality of the parameter space, we employed the Lanczos algorithm proposed by Ghorbani et al. \cite{ghorbani2019} to approximate the Hessian spectra and extract $\lambda_{\max}$. The resulting average $\lambda_{\max}$ trends for SGD (blue), SAM (red), and EISAM (purple) are presented in Fig. \ref{F3}. In practical terms, a lower $\lambda_{\max}$ reflects a flatter loss landscape, which in practice makes the model less sensitive to small data changes and improves its ability to generalize to new examples.

Our results demonstrate that both SAM and EISAM guide the model toward minima with lower curvature compared to SGD, but EISAM exhibits a more stable and decreasing trend in sharpness over epochs. This is evident from the $\lambda_{\max}$ trends, where SGD maintains a relatively high and stable level throughout training, indicating convergence to sharper minima. SAM shows some fluctuations in the early stage of training and gradually stabilizes in the later stage of training, suggesting potential instability. In contrast, EISAM displays a hign peak early on and a consistent decline toward low $\lambda_{\max}$ values in the later stages, implying a smoother trajectory to flatter regions. 

Specifically, during the entire training process, on the training set, the minimum observed $\lambda_{\max}$ for SGD is approximately 667.4, while for SAM and EISAM it is 50.0 and 77.1, respectively. On the test set, SGD is 797.8, SAM is 978.7, and EISAM is 314.8, indicating a significant reduction in curvature for EISAM on the test set. Additionally, the overall trend for EISAM shows a flatter profile with reduced variance, consistent with its ability to locate minima that enhance generalization. Notably, EISAM, as an efficient variant of SAM, preserves SAM’s capability to find flat minima while improving stability over epochs. The distinct trends in $\lambda_{\max}$ across SGD, SAM, and EISAM, which arise from EISAM’s optimized iterative strategy, underscore its strong potential for practical applications. These findings align with the hypothesis that flatter minima correlate with better generalization, providing empirical support for the effectiveness of sharpness-aware optimization techniques.

\begin{figure}[htbp]
    \centering
    \includegraphics[width=0.5
    \textwidth, angle=0, trim=0.5cm 0.2cm 1cm 0.2cm, clip]{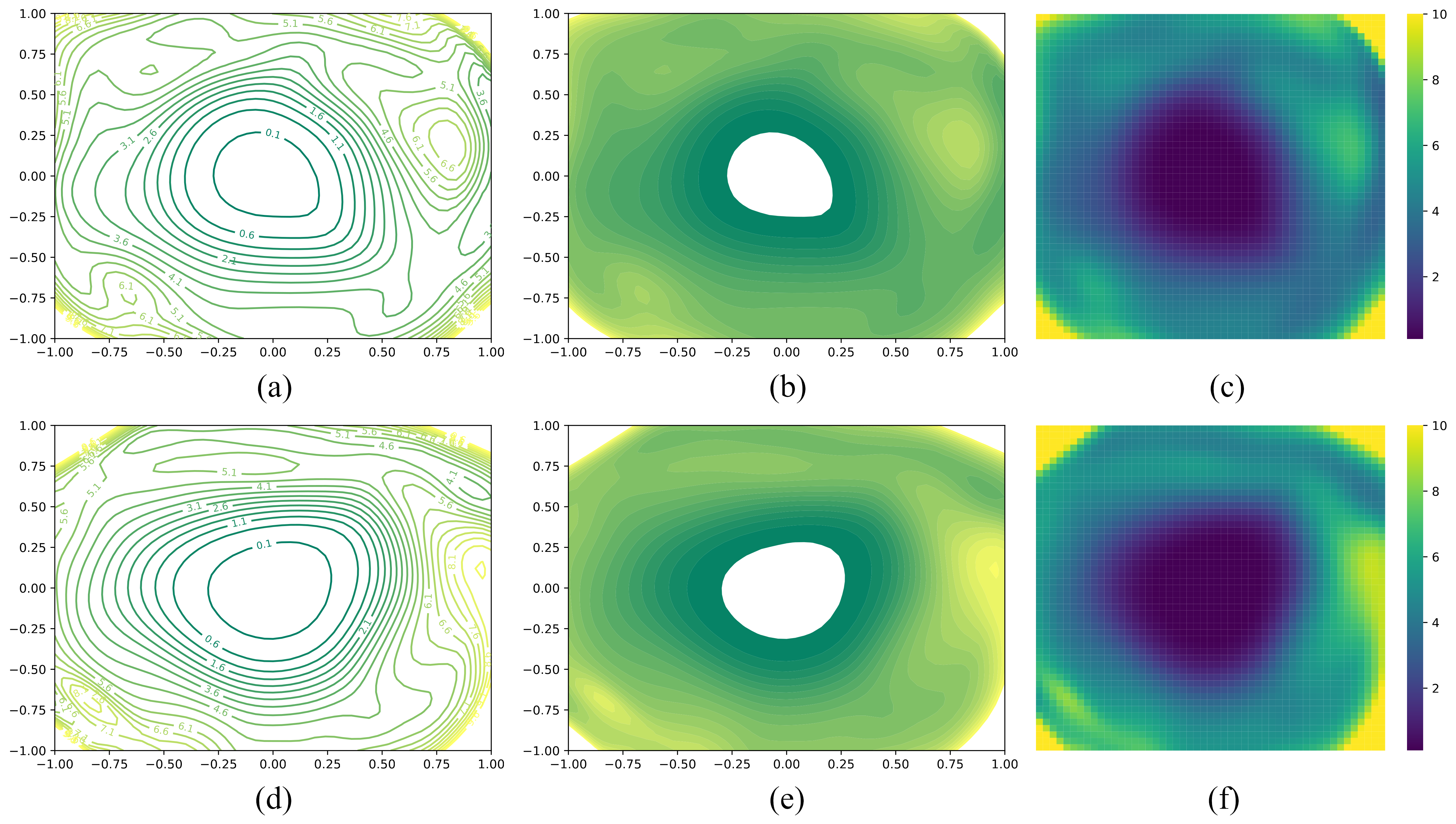}
    \caption{2D loss landscapes around the minima found by SAM and EISAM optimizers on the CIFAR-100 dataset using ResNet-50. The contour plots visualize the loss surface in a 2D projection, with color gradients indicating loss values (darker colors represent lower loss). Subfigures show different visualization styles for: (a) SAM (contour lines), (b) SAM (filled contours), (c) SAM (heatmap), (d) EISAM (contour lines), (e) EISAM (filled contours), (f) EISAM (heatmap).}
    \label{2D}
\end{figure}

Fig. \ref{2D} illustrates the 2D loss landscapes of SAM and EISAM through contour plots and heatmaps. SAM's landscape features steeper gradients near the center and a smaller low-loss region. In contrast, EISAM's loss landscape is flatter, with a larger circular low-loss area and more gradual gradient transitions. Overall, EISAM achieves a flatter loss landscape with a broader low-loss basin compared to SAM.

\section{EXCESS RISK ANALYSIS OF EISAM OPTIMIZER}
\label{sec4}

In this section, we provide a comprehensive theoretical analysis of the EISAM optimizer, demonstrating how its multi-step update rule guides model parameters toward regions of the loss landscape that enhance generalization. This analysis builds upon standard optimization theory and offers a robust foundation for EISAM's empirical success.

\subsection{Notations and Preliminaries}
\label{4.1}
Denoting $  \mathcal{X} \subset \mathbb{R}^p  $ and $  \mathcal{Y} \subset \mathbb{R}  $ as the feature and label spaces respectively, we consider an i.i.d. sample $  S  $ of size $  n  $: $S = \{z_i = (\mathbf{x}_i, y_i)\}_{i=1}^n,$
where each $  z_i  $ is drawn from an unknown distribution $  \mathfrak{D}  $ on $  \mathcal{Z} = \mathcal{X} \times \mathcal{Y}  $. A learning algorithm $  \mathcal{A}  $ maps the training set $  S  $ to a hypothesis $  h: \mathcal{X} \to \mathcal{Y}  $ parameterized by $  \mathbf{w}_{\mathcal{A},S}  $. We assume that $  \mathcal{A}  $ is symmetric with respect to $  S  $, i.e., its output does not depend on the ordering of the samples in $  S  $. For brevity, we omit the dependence on $  n  $.
We denote by $  \mathbb{E}_S[\cdot]  $ the expectation with respect to the random draw of the training set $  S  $, and by $  \mathbb{E}_{\mathfrak{D}}[\cdot]  $ (or $  \mathbb{E}_{z\sim\mathfrak{D}}[\cdot]  $) the expectation when $  z  $ is distributed according to $  \mathfrak{D}  $.
Let $  \ell: \mathcal{Y} \times \mathcal{Y} \to \mathbb{R}_+  $ be a nonnegative cost function. The per-sample loss of the hypothesis with respect to an example $z = (\mathbf{x},y)$ is given by
\begin{align*}
f(\mathbf{w}, z) \triangleq \ell(h(\mathbf{x}; \mathbf{w}), y),
\end{align*}
and the population risk (generalization error) is defined as
\begin{align*}
F_{\mathfrak{D}}(\mathbf{w}) = \mathbb{E}_{z \sim \mathfrak{D}}[f(\mathbf{w}, z)].
\end{align*}
Since the true distribution $\mathfrak{D}$ is inaccessible, we instead minimize the empirical risk over a finite training set $S$, namely, by using the empirical error instead
\begin{align*}
F_S(\mathbf{w}) = \frac{1}{n} \sum_{i=1}^n f(\mathbf{w}, z_i).
\end{align*}
The optimal minimizers for the empirical and population risks are denoted $\mathbf{w}_S^*$ and $\mathbf{w}_{\mathfrak{D}}^*$, respectively. Let $\mathbf{w}_{\mathcal{A}, S}$ represent the parameters learned by an algorithm $\mathcal{A}$ on dataset $S$. The expected excess risk, which measures the performance gap relative to the optimal population minimizer, is defined as:
\begin{align*}
\varepsilon_{\text{exc}} \triangleq \mathbb{E}[F_{\mathfrak{D}}(\mathbf{w}_{\mathcal{A}, S}) - F_{\mathfrak{D}}(\mathbf{w}_{\mathfrak{D}}^*)].
\end{align*}
This excess risk can be decomposed into three components: $\varepsilon_{\text{exc}} = \varepsilon_{\text{gen}} + \varepsilon_{\text{opt}} + \varepsilon_{\text{approx}}.$
Expected generalization error $\varepsilon_{\text{gen}} \triangleq \mathbb{E}[F_{\mathfrak{D}}(\mathbf{w}_{\mathcal{A}, S}) - F_S(\mathbf{w}_{\mathcal{A}, S})]$, which captures the gap between population and empirical risks; the optimization error $\varepsilon_{\text{opt}} \triangleq \mathbb{E}[F_S(\mathbf{w}_{\mathcal{A}, S}) - F_S(\mathbf{w}_S^*)]$, reflecting how well the algorithm minimizes the empirical risk; and $\mathbb{E}[F_{S}(\mathbf{w}_S^*) - F_{\mathfrak{D}}(\mathbf{w}_{\mathfrak{D}}^*)]=\mathbb{E}[F_{S}(\mathbf{w}_S^*) - F_{\mathfrak{D}}(\mathbf{w}_{S}^*)]+\mathbb{E}[F_{\mathfrak{D}}(\mathbf{w}_S^*) - F_{\mathfrak{D}}(\mathbf{w}_{\mathfrak{D}}^*)]$, the first term $\mathbb{E}[F_{S}(\mathbf{w}_S^*) - F_{\mathfrak{D}}(\mathbf{w}_{S}^*)]$ vanishes in over-parameterized models and is absorbed into the definition of approximation error $\varepsilon_{\text{approx}} \triangleq \mathbb{E}[F_{\mathfrak{D}}(\mathbf{w}_S^*) - F_{\mathfrak{D}}(\mathbf{w}_{\mathfrak{D}}^*)]$, which arises from using a finite sample and is typically negligible for large $n$ under standard statistical assumptions.
Our analysis focuses on bounding $\varepsilon_{\text{gen}}$ and $\varepsilon_{\text{opt}}$, assuming $\varepsilon_{\text{approx}}$ is small.

The following definitions formalize the key assumptions underpinning our derivations. These are standard in convex optimization literature and ensure the loss landscape is well-behaved. In what follows, $  \|\cdot\|  $ denotes the Euclidean norm for all vectors, and the subscript is omitted for brevity.
\begin{definition}[Strong Convexity]
Let $\mathcal{Z}$ be the sample space. The loss function $f(\mathbf{w}, z)$ is $\mu$-strongly convex if, for all $\mathbf{w}, \mathbf{v} \in \mathbb{R}^d$ and $z \in \mathcal{Z}$,
$$f(\mathbf{v}, z) \geq f(\mathbf{w}, z) + \nabla f(\mathbf{w}, z)^T (\mathbf{v} - \mathbf{w}) + \frac{\mu}{2} \|\mathbf{v} - \mathbf{w}\|^2,$$
where $\mu > 0$.
\end{definition}
\begin{definition}[Smoothness]
The loss function $f(\mathbf{w}, z)$ is $L$-smooth if, for all $\mathbf{w}, \mathbf{v} \in \mathbb{R}^d$ and $z \in \mathcal{Z}$,
$$\|\nabla f(\mathbf{v}, z) - \nabla f(\mathbf{w}, z)\| \leq L \|\mathbf{v} - \mathbf{w}\|,$$
where $L > 0$.
\end{definition}
\begin{definition}[Lipschitz Continuity]
The loss function $f(\mathbf{w}, z)$ is $G$-Lipschitz continuous if, for all $\mathbf{w}, \mathbf{v} \in \mathbb{R}^d$ and $z \in \mathcal{Z}$,
$$|f(\mathbf{v}, z) - f(\mathbf{w}, z)| \leq G \|\mathbf{v} - \mathbf{w}\|,$$
where $G > 0$.
\end{definition}
\begin{definition}[Hessian Lipschitz continuity]
The Hessian matrix $  \nabla^2 F_{B_t}  $ is Lipschitz continuous with constant $  K  $, i.e., $  \|\nabla^2 F_{B_t}(\mathbf{w}) - \nabla^2 F_{B_t}(\mathbf{v})\| \leq K \|\mathbf{w} - \mathbf{v}\|  $ for all $  \mathbf{w}, \mathbf{v}  $. Additionally, higher-order terms in the Taylor expansion are bounded such that the remainder $  R_t  $ satisfies $  \|R_t\| \leq \frac{K}{2} \delta_t^2  $ . 
\end{definition}
These assumptions imply that the Hessian matrix of the loss has eigenvalues bounded between $\mu$ and $L$, ensuring controlled curvature in the loss landscape.

The EISAM optimizer performs a three-step update at each iteration $t$, leveraging a mini-batch $B_t \subset S$ of size $|B_t|$:
\begin{enumerate}
\item \textbf{Compute Intermediate Point (Prediction Step)}:
$$\mathbf{y}_t = \mathbf{w}_t - s\nabla F_{B_t}(\mathbf{w}_t),$$
where $s > 0$ is the prediction step size, and $\nabla F_{B_t}(\mathbf{w}_t) = \frac{1}{|B_t|} \sum_{z \in B_t} \nabla f(\mathbf{w}_t, z)$ is the stochastic gradient on the mini-batch $B_t$. 
\item \textbf{Compute Perturbed Point (Sharpness-Aware Adjustment)}:
$$\mathbf{w}_t' = \mathbf{y}_t + \rho \frac{\nabla F_{B_t}(\mathbf{w}_t)}{\|\nabla F_{B_t}(\mathbf{w}_t)\|},$$
where $\rho > 0$ is the perturbation radius that controls the exploration of flat regions.
\item \textbf{Update Parameter}:
$$\mathbf{w}_{t+1} = \mathbf{y}_t - \eta \nabla F_{B_t}(\mathbf{w}_t'),$$
where $\eta > 0$ is the learning rate, and the update uses the gradient at the perturbed point to incorporate sharpness awareness.
\end{enumerate}
This structured update combines gradient descent with predictive and sharpness-aware mechanisms, balancing efficient minimization of the empirical risk with improved generalization.

\subsection{Stability Analysis}
\label{4.2}
Stability analysis is crucial for bounding the generalization error, as it quantifies how sensitive the learned parameters are to perturbations in the training data. We employ uniform stability \cite{Bousquet2002}, a concept that measures the maximum change in the loss for any test point when the training set is altered by one sample.
A learning algorithm $\mathcal{A}$ is $\varepsilon$-uniformly stable if, for any two training sets $S$ and $S'$ differing in at most one sample, and for all test points $z$,
\begin{align*}
\sup_{z} \mathbb{E} [\left| f(\mathbf{w}_{\mathcal{A}, S}, z) - f(\mathbf{w}_{\mathcal{A}, S'}, z) \right|] \leq \varepsilon.
\end{align*}
It follows that the expected generalization error satisfies $\varepsilon_{\text{gen}} \leq \varepsilon$.
To derive this bound for EISAM, let $\mathbf{w}_t$ and $\mathbf{v}_t$ denote the parameter trajectories when running EISAM on training sets $S$ and $S'$, respectively. Define the Euclidean distance between these trajectories as
\begin{align*}
\delta_t = \|\mathbf{w}_t - \mathbf{v}_t\|.
\end{align*}
Our analysis bounds the evolution of $\delta_t$ over $T$ iterations. We consider two cases based on whether the mini-batches $B_t$ (for $S$) and $B_t'$ (for $S'$) are identical, which occurs with probability $1 - \frac{1}{n}$ (since $S$ and $S'$ differ by one sample out of $n$).

\begin{lemma}[Parameter Trajectory Difference for Identical Mini-Batches]
\label{lem:identical}
Under Definitions 1--4, and using the co-coercivity lemma from strong convexity and smoothness (\cite{Hardt2016}),
\begin{align*}
&\langle \mathbf{w}_t' - \mathbf{v}_t', \nabla F_{B_t}(\mathbf{w}_t') - \nabla F_{B_t}(\mathbf{v}_t') \rangle \\
&\geq \frac{\mu L}{\mu + L} \left[(1 - s \mu) \delta_t + O(\delta_t^2)\right]^2.
\end{align*}
Consequently, for a learning rate satisfying $  \eta < \frac{2 (1 - s \mu)^2}{\mu + L}  $,
\begin{align*}
\delta_{t+1} \leq \left( 1 - \eta \frac{\mu L (1 - s \mu)^2}{\mu + L} \right) \delta_t + O(\delta_t^{3/2}),
\end{align*}
where the higher-order terms are negligible for sufficiently small $  \delta_t  $. 
\end{lemma}

In practical terms, this lemma shows that EISAM's prediction step reduces the divergence between parameter trajectories obtained from slightly different training sets. This strengthens the uniform stability of the algorithm and contributes to improved generalization.

\begin{lemma}[Parameter Trajectory Difference for Differing Mini-Batches]
\label{lem:differing}
Under the same assumptions, if the mini-batches differ,
\begin{align*}
\delta_{t+1} &\leq \left( 1 - \eta \frac{\mu L (1 - s \mu)^2}{\mu + L} \right) \delta_t + 2\eta G  \\
&+ O(\delta_t^{3/2} + \eta \rho \delta_t^2 / G+ \eta s K \delta_t^3).
\end{align*}
\end{lemma}
The proof follows from Lemma \ref{lem:identical} for the contraction term, with an additional $2\eta G$ arising from bounding the gradient norms $\|\nabla F_{B_t}(\mathbf{w}_t')\| \leq G$ and $\|\nabla F_{B_t'}(\mathbf{v}_t')\| \leq G$ using Lipschitz continuity. 

\begin{theorem}[Generalization Error Bound]
\label{thm:gen}
Under the assumptions of Lemmas \ref{lem:identical} and \ref{lem:differing}, including Hessian Lipschitz continuity with constant $  K  $ and bounded higher-order terms, the expected trajectory difference satisfies
\begin{align*}
\mathbb{E}[\delta_{t+1}] \leq (1 - \kappa) \mathbb{E}[\delta_t] + \frac{2\eta G}{n} + O\left( \mathbb{E}[\delta_t]^{3/2} \right),
\end{align*}
where $  \kappa = \eta \frac{\mu L (1 - s \mu)^2}{\mu + L}  $.
For sufficiently small initial $  \delta_0  $ and appropriate $  \eta  $ such that the higher-order term is controlled, the expected distance converges to
\begin{align*}
\mathbb{E}[\delta_T] \leq \frac{2 G (\mu + L)}{n \mu L (1 - s \mu)^2} + O\left( \left( \frac{\eta G (\mu + L)}{n \mu L (1 - s \mu)^2} \right)^{3/2} \right).
\end{align*}
By Lipschitz continuity, the generalization error is bounded by
\begin{align*}
\varepsilon_{\text{gen}} \leq \frac{2 G^2 (\mu + L)}{n \mu L (1 - s \mu)^2} + O\left( \left( \frac{\eta G^2 (\mu + L)}{n \mu L (1 - s \mu)^2} \right)^{3/2} \right).
\end{align*}
\end{theorem}
This bound shows that EISAM's prediction step tightens the leading term of the generalization error by a factor involving $  (1 - s \mu)^2  $, improving upon standard SAM bounds, while the higher-order term reflects the approximation error controlled by the Hessian Lipschitz constant $  K  $, small perturbation $\rho$, and prediction step $s$.

\subsection{Convergence Analysis}
\label{4.3}
Convergence analysis focuses on bounding the optimization error $\varepsilon_{\text{opt}}$, which quantifies how close EISAM gets to the empirical minimizer $\mathbf{w}_S^*$. We begin by examining the progress made in a single iteration.
\begin{lemma}[One-Step Progress]
\label{lem:progress}
Under Definitions 1--4. The one-step update satisfies
\begin{align*}
&F_{B_t}(\mathbf{w}_{t+1}) - F_{B_t}(\mathbf{w}^*) \\
&\leq (1 - 2\eta\mu (1 - s \mu)) (F_{B_t}(\mathbf{w}_t) - F_{B_t}(\mathbf{w}^*)) + \frac{\eta^2 L G^2}{2} \\
&+ O(\eta s^2 G^3 + \eta \rho G^2).
\end{align*}
\end{lemma}

The bound features a contraction factor of $(1 - s \mu)$ arising from the prediction step. The perturbation and higher-order terms introduce an additional error of order $O(\eta s^2 G^3 + \eta \rho G^2)$, which remains controllable for sufficiently small $s$ and $\rho$.

\begin{theorem}[Optimization Error Bound]
\label{thm:opt}
Under the assumptions of Lemma \ref{lem:progress}, the expected suboptimality after sufficient iterations satisfies
\begin{align*}
\varepsilon_{\text{opt}} \leq \frac{\eta L G^2}{4 \mu (1 - s \mu)} + O\left( \eta s^2 G^3 + \eta \rho G^2 \right),
\end{align*}
assuming the process converges and higher-order terms are controlled by sufficiently small $s$ and $\rho$.
\end{theorem}
The leading term shows that the prediction step leads to faster convergence by introducing the factor $(1 - s \mu)$ in the rate. The higher-order terms involving $s$ and $\rho$ are controllable for sufficiently small values of these parameters.

\subsection{Expected Excess Risk}
\label{4.4}
Combining the generalization and optimization bounds provides an overall measure of EISAM's performance.
\begin{theorem}[Expected Excess Risk Bound]
\label{thm:exc}
Under the assumptions of Theorems \ref{thm:gen} and \ref{thm:opt}, with the learning rate constraint \(\eta \leq \frac{2 (1 - s \mu)^2}{\mu + L}\), the expected excess risk of the output $\mathbf{w}_{\mathcal A,S}$ is bounded by
\begin{align*}
\varepsilon_{\text{exc}}&\leq \varepsilon_{\text{opt}}+\varepsilon_{\text{gen}}\leq \frac{2 G^2 (\mu + L)}{n \mu L (1 - s \mu)^2} + \frac{\eta L G^2}{4 \mu (1 - s \mu)} \\
&+ O\left( \left( \frac{\eta G^2 (\mu + L)}{n \mu L (1 - s \mu)^2} \right)^{3/2} + \eta s^2 G^3 + \eta \rho G^2 \right),
\end{align*}
\end{theorem}
The trade-off between generalization and optimization can be observed from the leading terms: the generalization bound contains the factor $(1 - s \mu)^2$ in the denominator (tighter than the standard SAM case where the factor is 1), while the optimization term is affected through $(1 - s \mu)$. The perturbation $\rho$ primarily affects higher-order terms and should be chosen small to keep errors controllable. In practice, the optimal balance between $s$ and $\rho$ can be tuned empirically.

This implies that a moderate increase in the prediction step size $s$ can significantly tighten the generalization bound with only a mild effect on the optimization rate. Moreover, since $\rho$ only influences higher-order terms, EISAM exhibits reduced sensitivity to the choice of perturbation radius compared to vanilla SAM, thereby simplifying hyperparameter tuning in practice.

\subsection{Nonconvex Extension of Excess Risk Analysis}

In the main analysis (strong convexity setting), we provided tight bounds for tractability. In this section, we extend the analysis to the nonconvex case by relaxing strong convexity to L-smoothness and bounded stochastic gradient variance.

\begin{definition}[L-Smoothness and Bounded Variance]
\label{def5}
The population loss $F(\mathbf{w})$ is L-smooth: $\|\nabla F(\mathbf{w}) - \nabla F(\mathbf{v})\| \leq L \|\mathbf{w} - \mathbf{v}\|$ for all $\mathbf{w}, \mathbf{v} \in \mathbb{R}^d$. The stochastic gradients satisfy $\mathbb{E}[\|\nabla f(\mathbf{w}; z) - \nabla F(\mathbf{w})\|^2] \leq \sigma^2$, where $z \sim \mathfrak{D}$.
\end{definition}

\begin{theorem}[Nonconvex Expected Excess Risk Bound]
\label{thm:Nonconvex exc}
Under Definition \ref{def5}, with learning rate $\eta$, perturbation parameters $\rho, s > 0$ (assuming $\eta, s, \rho$ small enough so that $1 - s L - \rho L > 0$), after $T$ iterations, the expected excess risk for averaged parameters $\bar{\mathbf{w}}_T = \frac{1}{T} \sum_{t=1}^T \mathbf{w}_t$ satisfies:
\begin{align*}
&\mathbb{E}[F(\bar{\mathbf{w}}_T) - F(\mathbf{w}^*)] \leq \mathcal{O}\left( \frac{F(\mathbf{w}_1) - F^*}{\eta T (1 - s L - \rho L)} \right)\\
&+ \mathcal{O}\left(\eta L \sigma^2+s^2 L \sigma^2 + \rho^2 L + \sigma \sqrt{\frac{d \log T}{n}} \right),
\end{align*}
where $F^* = \inf_{\mathbf{w}} F(\mathbf{w})$, $d$ is the parameter dimension, and $n$ is the sample size.
\end{theorem}
In nonconvex settings, this sublinear $\mathcal{O}(1/\sqrt{T})$-level bound decomposes the expected excess risk into an optimization term of order $\mathcal{O}\left( \frac{F(\mathbf{w}_1) - F^*}{\eta T (1 - s L - \rho L)} \right)$, variance terms of order $O(\eta L \sigma^2 + s^2 L \sigma^2 + \rho^2 L)$, and a generalization term of order $\mathcal{O}(\sigma \sqrt{d \log T / n})$.

The prediction step size $s$ and perturbation radius $\rho$ appear explicitly in the convergence rate via the denominator factor $1 - s L - \rho L$ and in the higher-order variance terms via the coefficients $s^2$ and $\rho^2$. This structure reveals a direct trade-off: larger values of $s$ or $\rho$ can accelerate the leading convergence rate but increase the variance contribution to the bound, while smaller values lead to more stable optimization at the potential cost of slower convergence.

\section{Experiments}
\label{sec5}

\begin{table*}[h]
    \centering
    \caption{Results on CIFAR-10. We run each model with three different random seeds and report the mean test accuracy (\%) along with the standard deviation. Text marked as bold indicates the best result.}
    \label{T1}
    \begin{tabular*}{\textwidth}{@{\extracolsep{\fill}} l c c c c c c}
        \toprule
        Architecture & SGD & SAM & ASAM & GSAM & FSAM & EISAM (Ours) \\
        \midrule
        ResNet-18 & $96.41_{\pm0.07}$ & $96.76_{\pm0.09}$ & $96.48_{\pm0.07}$ & $96.65_{\pm0.16}$ & $96.60_{\pm0.03}$ & $\bf96.84_{\pm0.09}$ \\
        ResNet-50 & $96.56_{\pm0.09}$ & $96.86_{\pm0.05}$ & $96.58_{\pm0.08}$ & $97.11_{\pm0.12}$ & $97.08_{\pm0.24}$ &  $\bf97.13_{\pm0.06}$ \\
        ResNet-101 & $96.83_{\pm0.13}$ & $97.27_{\pm0.06}$ & $97.14_{\pm0.14}$ & $97.31_{\pm0.12}$ & $97.26_{\pm0.14}$ & $\bf97.48_{\pm0.12}$ \\
        WideResNet-28-10 & $96.86_{\pm0.08}$ & $97.30_{\pm0.04}$ & $97.28_{\pm0.07}$ & $97.49_{\pm0.08}$ & $97.55_{\pm0.08}$ & $\bf97.63_{\pm0.12}$ \\
        PyramidNet-110 & $97.27_{\pm0.08}$ & $97.69_{\pm0.03}$ & $97.50_{\pm0.08}$ & $97.76_{\pm0.03}$ & $97.93_{\pm0.10}$ & $\bf98.01_{\pm0.07}$\\
        \bottomrule
    \end{tabular*}
\end{table*}
\begin{table*}[h]
    \centering
    \caption{Results on CIFAR-100. We run each model with three different random seeds and report the mean test accuracy (\%) along with the standard deviation. Text marked as bold indicates the best result.}
    \label{T2}
    \begin{tabular*}{\textwidth}{@{\extracolsep{\fill}} l c c c c c c}
        \toprule
        Architecture & SGD & SAM & ASAM & GSAM & FSAM &  EISAM (Ours) \\
        \midrule
        ResNet-18 & $80.92_{\pm0.17}$ & $81.91_{\pm0.15}$ & $81.88_{\pm0.12}$ & $81.39_{\pm0.19}$ & $82.02_{\pm0.17}$ & $\bf82.22_{\pm0.14}$ \\
        ResNet-50 & $81.28_{\pm0.45}$ & $83.09_{\pm0.37}$ & $82.32_{\pm0.29}$ & $82.06_{\pm0.36}$ & $83.23_{\pm0.46}$ & $\bf83.64_{\pm0.17}$ \\
        ResNet-101 & $82.50_{\pm0.20}$ & $83.62_{\pm0.11}$ & $83.00_{\pm0.14}$ & $83.43_{\pm0.30}$ & $84.21_{\pm0.30}$ & $\bf84.44_{\pm0.13}$ \\
        WideResNet-28-10 & $83.31_{\pm0.41}$ & $85.23_{\pm0.01}$ & $83.79_{\pm0.12}$ & $84.05_{\pm0.11}$ & $85.16_{\pm0.06}$ &  $\bf85.85_{\pm0.11}$ \\
        PyramidNet-110 & $83.98_{\pm0.09}$ & $86.61_{\pm0.27}$ & $84.90_{\pm0.13}$ & $85.02_{\pm0.13}$ & $85.91_{\pm0.17}$ & $\bf86.92_{\pm0.14}$\\
        \bottomrule
    \end{tabular*}
\end{table*}

In this study, we conducted a series of experiments to evaluate the performance of the EISAM optimizer across various deep learning tasks, including image classification, natural language processing, object detection, and image segmentation. The primary objective was to assess EISAM's effectiveness in improving generalization and training efficiency compared to established optimizers such as SGD, SAM \cite{Foret2021}, ASAM \cite{Kwon2021}, GSAM \cite{zhuang2022surrogate}, FSAM \cite{du2024friendly} and Adam \cite{Kingma2015}. 
Specifically, the compared optimizers are as follows:\\
\textbf{Adam}: A widely adopted adaptive optimizer known for its efficiency in training deep neural networks, with two key hyperparameters: learning rate (lr) and weight decay (wd).\\
\textbf{SAM}: A sharpness-aware optimization method that seeks flatter minima in the loss landscape, with three hyperparameters: learning rate (lr), weight decay (wd), and perturbation radius $  \rho  $.\\
\textbf{ASAM}: An adaptive variant of SAM that dynamically adjusts the perturbation to further enhance generalization, with three hyperparameters: learning rate (lr), weight decay (wd), and perturbation radius $  \rho  $.\\
\textbf{GSAM}: A generalized sharpness-aware method that minimizes both the perturbed loss and a surrogate gap via orthogonal ascent, with four hyperparameters: learning rate (lr), weight decay (wd), perturbation radius $  \rho  $, and an ascent strength parameter $  \alpha  $.\\
\textbf{FSAM}: A friendly variant of SAM that employs Fisher-mask-based sparse perturbations and explicit parameter restoration, with four hyperparameters: learning rate (lr), weight decay (wd), perturbation radius $  \rho  $, and EMA decay factor $  \lambda  $.\\
\textbf{EISAM (ours)}: An extragradient-inspired extension of SAM that introduces an additional prediction step size $  s  $ to probe the loss landscape geometry, with four hyperparameters: learning rate (lr), weight decay (wd), perturbation radius $  \rho  $, and prediction step size $  s  $.

Our experiments were designed to test EISAM under diverse conditions, including different datasets, model architectures, and hyperparameter settings, with a particular focus on its adaptability and robustness.

For image classification, we utilized benchmark datasets such as CIFAR-10, CIFAR-100, and ImageNet-1K, employing a range of model architectures including ResNet \cite{He2016}, WideResNet \cite{Zagoruyko2016}, PyramidNet \cite{Han2017}, and Vision Transformers (ViT) \cite{dosovitskiy2021vit}. The experiments incorporated data augmentation techniques CutMix to enhance model robustness. In the domain of object detection, we leveraged the COCO 2017 and LVIS v1.0 datasets with the Faster R-CNN model \cite{ren2017fasterrcnn}, while for image segmentation, the ISIC2018 dataset was used with U-Net architectures \cite{ronneberger2015unet}. Hyperparameter sensitivity was also thoroughly analyzed to understand the impact of EISAM's additional parameters, $\rho$ (perturbation radius) and $s$ (prediction step size), on model performance.

To ensure the reliability and reproducibility of our results, all experiments were conducted with three random seeds, and performance metrics were averaged across these runs. The training process for each task was standardized, with consistent learning rate schedules, batch sizes, and early stopping criteria when applicable. This rigorous experimental framework allows for a comprehensive evaluation of EISAM's capabilities and provides insights into its potential as a versatile optimization tool for deep learning applications. For a comprehensive comparison of all experiments, see Fig. \ref{huiz} in Appendix \ref{all}.

\subsection{Image Classification}

For CIFAR-10 and CIFAR-100, to address overfitting in the absence of complex augmentation, CutMix was employed. CutMix combines two training images by cutting and pasting patches, which has been proven to enhance model robustness and generalization. On ImageNet-1K, no data augmentation was used to maintain experimental consistency.

All optimizers on CIFAR-10 and CIFAR-100 were trained for 200 epochs with a cosine learning rate decay schedule. This strategy gradually reduces the learning rate from its initial value to zero, promoting stable convergence. To determine the optimal hyperparameter configuration, a grid search was conducted over the following ranges:
\begin{itemize}
    \item Learning rate: \{0.01, 0.05, 0.1\}
    \item Mini-batch size: 128
    \item Weight decay: \{5.0e-4, 2.0e-3, 1.0e-3\}
    \item Perturbation radius $\rho$ : \{0.05, 0.1, 0.2\} for SAM, GSAM, FSAM and EISAM, \{0.5, 1, 2\} for ASAM
    \item Prediction step size $s$: \{1.0e-3, 5.0e-3\}
\end{itemize}

For ImageNet-1K, we adopt the same settings for both models: a mini-batch size of 256 and a cosine learning rate decay schedule to gradually reduce the learning rate from its initial value to zero, promoting stable convergence. The default base optimizer is SGD with an initial learning rate of 0.1 and weight decay of 1.0e-4. For SAM-family optimizers (including SAM and its variants), the perturbation radius $\rho$ is searched within the range \{0.01, 0.05, 0.1\}. For our proposed EISAM optimizer, the newly introduced hyperparameter $s$ is fixed as 1.0e-3.

For each hyperparameter combination, the experiment was run with 3 different random seeds to ensure statistical reliability. The average test accuracy and standard deviation across these runs are reported. Table \ref{T1} and Table \ref{T2} show the average test accuracies and standard deviations for different model architectures trained using EISAM as well as the baseline optimizers on CIFAR-10 and CIFAR-100. Table \ref{T3} shows the average test accuracies and standard deviations of the ResNet18 and ResNet50 models trained using EISAM as well as the baseline optimizers on ImageNet-1K.

\begin{figure*}[htbp]
    \centering
    \includegraphics[width=1\textwidth, angle=0, trim=0.2cm 0.1cm 0.2cm 0.2cm, clip]{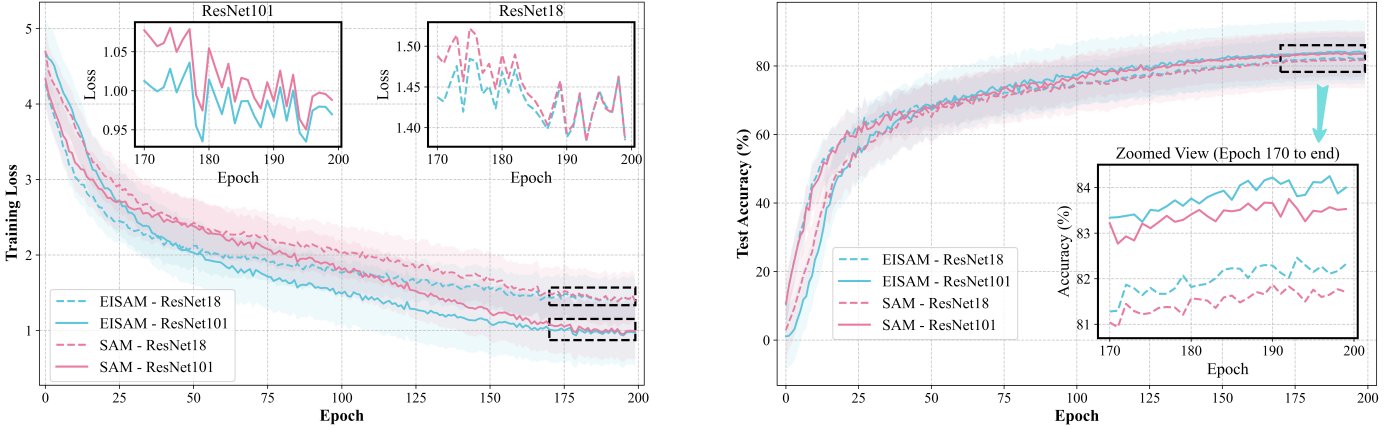}
    \caption{The training loss and test accuracy of ResNet18 and ResNet101 on CIFAR-100 with CutMix Augmentation.}
    \label{F5}
\end{figure*}

\begin{table*}[h]
    \centering
    \caption{Results on ImageNet-1K. We run each model with three different random seeds and report the mean test accuracy (\%) along with the standard deviation. Text marked as bold indicates the best result.}
    \label{T3}
    \begin{tabular*}{\textwidth}{@{\extracolsep{\fill}} l c c c c c c c}
        \toprule
        Architecture & SGD & SAM & ASAM & GSAM & FSAM & EISAM (Ours) \\
        \midrule
        ResNet-18 & $70.56_{\pm0.03}$ & $70.74_{\pm0.22}$ & $70.69_{\pm0.17}$ & $70.66_{\pm0.11}$ & $70.24_{\pm0.36}$ & $\bf70.89_{\pm0.15}$ \\
        ResNet-50 & $77.09_{\pm0.12}$ & $77.81_{\pm0.04}$ & $77.63_{\pm0.18}$ & $77.35_{\pm0.12}$ & $77.20_{\pm0.14}$ & $\bf78.01_{\pm0.07}$\\
        \bottomrule
    \end{tabular*}
\end{table*}

Fig. \ref{F5} illustrate the training loss and test accuracy of EISAM versus baseline optimizers on CIFAR-10 and CIFAR-100 using CutMix augmentation. EISAM achieves higher accuracy and lower variance across most architectures, demonstrating its superiority in mitigating overfitting. 

 The training loss trends show that all models start with a loss around 4 and drop sharply within the first 50 epochs. EISAM-based models reach lower final training losses, demonstrating their ability to effectively minimize the training objective. The EISAM optimizer's training loss decreases faster, indicating that it can converge faster. EISAM models, on the other hand, achieve higher test accuracy and exhibit better generalization performance, as evidenced by their consistently superior results across the evaluated datasets and architectures. These results demonstrate that EISAM’s enhanced method strikes the best balance between training efficiency and generalization, making it highly effective for deep learning tasks.

Furthermore, we assessed EISAM's performance using the ViT \cite{dosovitskiy2021vit} architecture on CIFAR-10. Experiments were conducted on two ViT variants, ViT-S-8 and ViT-S-16, which differ in patch size (8 and 16, respectively). The patch size controls the division of input images into segments processed by the transformer. For a comprehensive evaluation, each variant was tested under two training configurations:
\\
\textbf{Basic Configuration}: This setup involved standard training without any data augmentation techniques, serving as a baseline to assess the optimizer's performance in a vanilla setting.
\\
\textbf{CutMix Configuration}: This configuration incorporated the CutMix data augmentation technique, which mixes portions of different images and their labels to enhance model robustness and generalization.

The ViT models were trained for 200 epochs, with all optimizers incorporated into the training loop. The experimental results are presented in Table \ref{T4}, showcasing the mean test accuracy and standard deviation for each model and optimizer combination.

\begin{table*}[h]
    \centering
    \caption{Results on CIFAR-10. We run each model with three different random seeds and report the mean test accuracy (\%) along with the standard deviation. Text marked as bold indicates the best result.}
    \label{T4}
    \begin{tabular*}{\textwidth}{@{\extracolsep{\fill}} l c c c c c c}
        \toprule
        Architecture & Adam & SAM & ASAM & GSAM & FSAM & EISAM (Ours) \\
        \midrule
        ViT-S-8(Basic) & $78.50_{\pm0.33}$ & $78.32_{\pm0.16}$ & $78.16_{\pm0.14}$ & $78.35_{\pm0.09}$ & $77.93_{\pm0.34}$ & $\bf78.66_{\pm0.12}$ \\
        ViT-S-8(Cutmix) & $86.30_{\pm0.25}$ & $86.26_{\pm0.19}$ & $86.21_{\pm0.24}$ & $84.52_{\pm0.24}$ & $83.97_{\pm0.10}$ & $\bf86.42_{\pm0.05}$ \\
        ViT-S-16(Basic) & $68.33_{\pm0.44}$ & $68.68_{\pm0.11}$ & $68.77_{\pm0.16}$ & $68.60_{\pm0.54}$ & $\bf69.04_{\pm0.55}$ & $68.88_{\pm0.14}$ \\
        ViT-S-16(Cutmix) & $75.16_{\pm0.20}$ & $76.54_{\pm0.19}$ & $77.12_{\pm0.13}$ & $74.42_{\pm0.20}$ & $73.80_{\pm0.50}$ & $\bf77.46_{\pm0.15}$ \\
        \bottomrule
    \end{tabular*}
\end{table*}

EISAM's Superiority: In all configurations, EISAM consistently outperformed SAM, Adam, and GSAM, particularly with the ViT-S-8 model, although it was occasionally slightly outperformed by FSAM. This suggests that EISAM's enhanced iterative strategy effectively leverages the transformer's architecture, leading to improved generalization on CIFAR-10.
Impact of CutMix: The addition of CutMix significantly boosted performance for all optimizers, where EISAM benefited the most in the ViT-S-8 setup. This indicates that EISAM synergizes well with data augmentation, amplifying its optimization capabilities.

The experiments demonstrate that EISAM is a highly effective optimizer for Vision Transformer models, particularly with smaller patch sizes (e.g., ViT-S-8) and when combined with CutMix augmentation. Its ability to consistently outperform other optimizers in most scenarios underscores its potential as a robust optimization tool for transformer-based image classification tasks. These findings provide valuable insights for advancing transformer-based models in computer vision applications.

\subsection{Natural Language Processing}
In this section, we present the experimental setup and results of training the T5 model \cite{raffel2020t5} on the BOOLQ dataset. We searched for the best hyperparameter combination using a hyperparameter grid, and the best results are averaged over three random seeds to evaluate their performance in terms of best validation accuracy and best test accuracy.

\begin{table}[h]
    \centering
    \caption{Results on BOOLQ using T5. We run each model with three different random seeds and report the mean validation accuracy (\%) and test accuracy (\%) along with the standard deviation. Text marked as bold indicates the best result.}
    \label{T5}
    \begin{tabular*}{\linewidth}{@{\extracolsep{\fill}} l c c}
        \toprule
        Optimizer & Validation & Test \\
        \midrule
        Adam & $76.17\pm0.38$ & $75.18\pm1.32$ \\
        SAM & $77.14\pm0.50$ & $76.02\pm0.80$ \\
        ASAM & $77.16\pm0.43$ & $78.19\pm0.76$ \\
        GSAM & $76.94\pm0.93$ & $78.32\pm0.63$ \\
        FSAM & $76.19\pm0.83$ & $78.41\pm0.56$ \\
        EISAM(Ours) & $\bf77.19\pm0.36$ & $\bf78.64\pm0.81$ \\
        \bottomrule
    \end{tabular*}
\end{table}

We used the BOOLQ (Boolean Questions) dataset, a question-answering benchmark consisting of yes/no questions paired with short passages. The dataset was split into training, validation, and test sets with ratios of 70\%, 10\%, and 20\%, respectively. To enhance robustness, we applied data augmentation via synonym replacement during training. The model employed is the t5-small variant of the T5 (Transfer Text-to-Text Transformer) architecture, a widely-used text-to-text framework suitable for question-answering tasks like BOOLQ. We compared five optimizers:

To identify the optimal settings for each optimizer, we conducted a grid search over the following hyperparameter ranges:
\begin{itemize}
\item Learning rate: \{1e-4, 5e-4, 1e-3\}
\item Weight decay: \{1e-5, 5e-5, 5e-4\}
\item Perturbation radius: \{1e-3, 5e-3, 0.1, 0.05, 0.01\} 
\item GSAM hyperparameter $\alpha$: \{0.02, 0.05, 0.2\}
\item FSAM hyperparameter $\lambda$: \{0.6, 0.9, 0.95\}
\item EISAM hyperparameter $s$: \{5e-5, 1e-4, 5e-4\}
\end{itemize}
    
In this study, we evaluated the performance of each optimizer configuration under its best hyperparameters, by calculating the average best validation accuracy and best test accuracy. Training was performed for 30 epochs, with early stopping applied if validation accuracy did not improve for 5 consecutive epochs. The learning rate was scheduled with a linear warmup followed by cosine decay. Performance was assessed using:
\\
\textbf{Best validation accuracy:} The highest accuracy on the validation set during training.
\\
\textbf{Best test accuracy:} The accuracy on the test set using the model checkpoint with the best validation performance.

The comparative experimental results of different equipment combinations are summarized in Table \ref{T5}, which presents the best validation and test accuracies for each configuration.

As the baseline, AdamW achieved solid performance, with an average best validation accuracy of 76.17\% and test accuracy of 75.18\%. However, it was outperformed by the sharpness-aware optimizers, suggesting limitations in its ability to generalize under the given settings. However, SAM shows limited generalization improvement on natural language tasks, while SAM variants can further enhance generalization on such tasks, yielding 77.16\% validation accuracy and 78.41\% test accuracy. This indicates that sharpness-aware optimization enhances generalization on BOOLQ. EISAM outperformed both optimizers, achieving 77.19\% validation accuracy and 78.64\% test accuracy. The addition of the prediction step size (s) likely enabled EISAM to better navigate the optimization landscape, balancing exploration and exploitation more effectively.

These results highlight EISAM's potential as an effective optimizer for deep learning tasks requiring high precision and generalization. 

\subsection{Object Detection}
This section delineates a systematic experimental investigation designed to scrutinize the efficacy of the EISAM optimizer within the domain of object detection. The overarching aim of this study is to juxtapose EISAM’s optimization capabilities against those of established methodologies. All experiments adhere to default hyperparameter configurations to ensure consistency and reproducibility.

The evaluation first leverages the COCO 2017 dataset \cite{lin2014coco}, an authoritative benchmark renowned for its complexity and diversity, encompassing over 100,000 images spanning 80 object categories. To provide a more stringent test of optimization robustness and generalization from scratch, we train the Faster R-CNN architecture with a ResNet-50 backbone using completely randomly initialized weights (rather than any pre-trained initialization). This setup eliminates the influence of pre-trained features and directly assesses the optimizer’s ability to discover flat minima in a challenging detection landscape. In Table \ref{coco}, we report the standard COCO metrics including AP (average precision), $\mathrm{AP}_{50}$, $\mathrm{AP}_{75}$, $\mathrm{AP}_{S}$, $\mathrm{AP}_{M}$, and $\mathrm{AP}_{L}$.

To further validate EISAM on large-vocabulary long-tailed detection, we additionally conduct full-parameter fine-tuning experiments on the LVIS v1.0 dataset \cite{gupta2019lvis} using the pre-trained Faster R-CNN with ResNet-50 FPN backbone. LVIS v1.0 contains 1203 categories with a severe long-tailed distribution, thereby stressing the optimizer’s capacity to balance performance across rare, common, and frequent classes. In Table \ref{lvis}, we report the comprehensive metrics including AP, $\mathrm{AP}_{50}$, $\mathrm{AP}_{75}$, $\mathrm{AP}_{S}$, $\mathrm{AP}_{M}$, $\mathrm{AP}_{L}$, $\mathrm{AP}_{R}$, $\mathrm{AP}_{C}$, and $\mathrm{AP}_{F}$.

\begin{table*}[htbp]
    \centering
    \caption{Object detection results of different optimizers on COCO validation set.}
    \label{coco}
    \small
    \setlength{\tabcolsep}{12pt} 
    \begin{tabular}{l c c c c c c}
        \toprule
        Optimizer & AP & $\mathrm{AP}_{50}$ & $\mathrm{AP}_{75}$ & $\mathrm{AP}_{S}$ & $\mathrm{AP}_{M}$ & $\mathrm{AP}_{L}$\\
        \midrule
        SGD & $22.23$ & $39.08$ & $22.59$ & $12.38$ & $22.97$ & $30.70$ \\
        SAM & $27.53$ & $45.51$ & $\bf29.04$ & $\bf15.86$ & $29.21$ & $36.03$ \\
        ASAM & $23.37$ & $40.52$ & $23.86$ & $12.49$ & $24.23$ & $32.60$ \\
        GSAM & $21.90$ & $38.59$ & $22.24$ & $11.82$ & $22.74$ & $30.52$ \\
        FSAM & $27.37$ & $45.59$ & $28.99$ & $15.78$ & $29.13$ & $35.38$ \\
        EISAM (Ours) & $\bf27.65$ & $\bf45.63$ & $29.01$ & $15.85$ & $\bf29.37$  & $\bf36.12$\\
        \bottomrule
    \end{tabular}
\end{table*}

These two complementary experimental settings, training from scratch on COCO 2017 and large-vocabulary fine-tuning on LVIS v1.0, establish a robust foundation for evaluating the EISAM optimizer’s superiority in object detection tasks, consistently demonstrating improved generalization and training efficiency over baseline optimizers (SGD, Adam, and SAM).

\begin{table*}[htbp]
    \centering
    \caption{Object detection results of different optimizers on LVIS validation set.}
    \label{lvis}
    \small
    \begin{tabular}{l c c c c c c c c c}
        \toprule
        Optimizer & AP & $\mathrm{AP}_{50}$ & $\mathrm{AP}_{75}$ & $\mathrm{AP}_{S}$ & $\mathrm{AP}_{M}$ & $\mathrm{AP}_{L}$ & $\mathrm{AP}_{R}$ & $\mathrm{AP}_{C}$ & $\mathrm{AP}_{F}$\\
        \midrule
        SGD & $10.37$ & $18.00$ & $10.78$ & $7.35$ & $14.60$ & $20.03$ & $\bf0.04$ & $5.25$ & $21.87$ \\
        SAM & $11.74$ & $19.10$ & $\bf12.51$ & $8.02$ & $16.55$ & $\bf22.05$ & $0.004$ & $5.45$ & $\bf24.38$\\
        ASAM & $10.28$ & $17.84$ & $10.66$ & $7.29$ & $14.52$ & $19.23$ & $0.03$ & $4.82$ & $22.22$\\
        GSAM & $10.38$ & $17.94$ & $10.69$ & $7.17$ & $14.54$ & $19.21$ & $0.02$ & $4.89$ & $22.15$\\
        FSAM & $11.71$ & $19.15$ & $12.43$ & $\bf8.05$ & $16.54$ & $21.76$ & $0.008$ & $5.55$ & $24.30$\\
        EISAM (Ours) & $\bf11.76$ & $\bf19.23$ & $12.47$ & $\bf8.05$ & $\bf16.60$  & $\bf21.99$ & $0.005$ & $\bf5.75$ & $24.26$\\
        \bottomrule
    \end{tabular}
\end{table*}

\begin{figure*}[htbp]
\centering
\includegraphics[width=1\textwidth, angle=0, trim=0.1cm 0.1cm 0.2cm 0.1cm, clip]{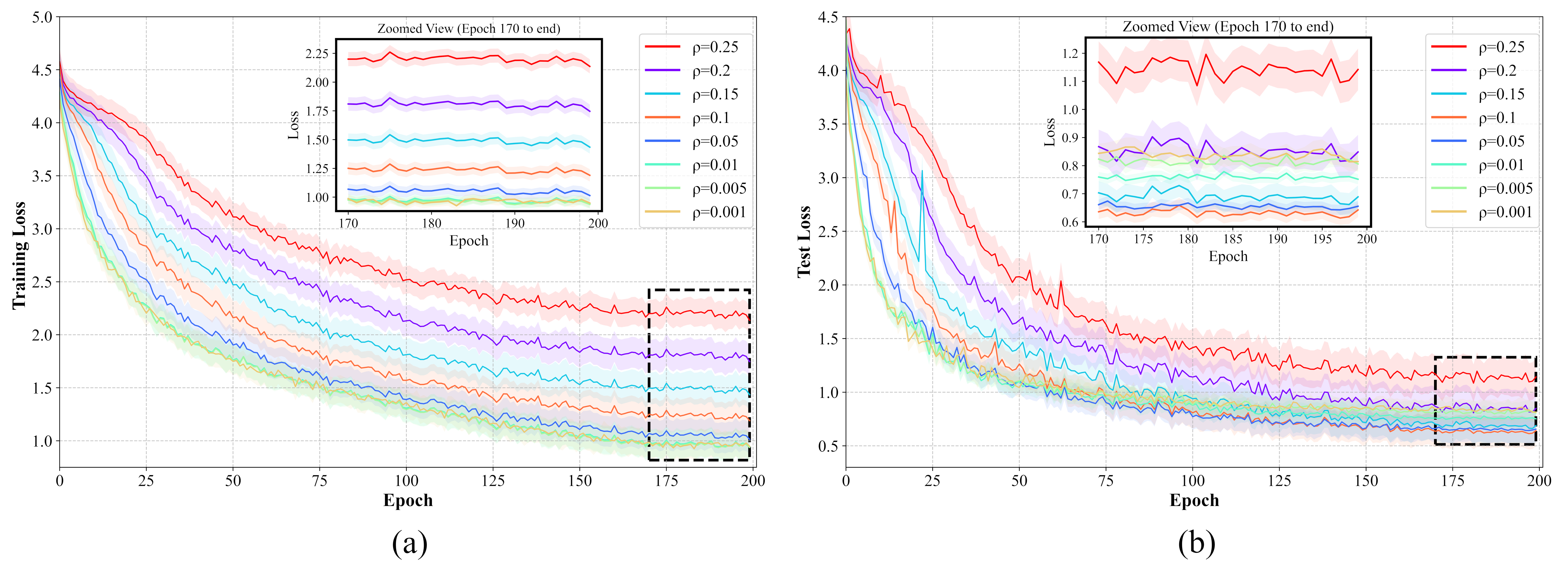}
\caption{EISAM hyperparameter sensitivity of $\rho$: (a) Training loss, (b) Test loss. In this set of experiments, we train a ResNet-50 model to classify the CIFAR-100 dataset. All models are trained with the same initial learning rate of 0.01, mini-batch size of 128, and weight decay of 1.0e-3 for 200 epochs. To explore the effect of perturbation radius $\rho$, we sweep $\rho$ across a range of values: [0.001, 0.005, 0.01, 0.05, 0.1, 0.15, 0.2, 0.25].}
\label{rho2}
\end{figure*}
\begin{figure*}[htbp]
\centering
\includegraphics[width=1\textwidth, angle=0, trim=0.2cm 0.1cm 0.2cm 0.1cm, clip]{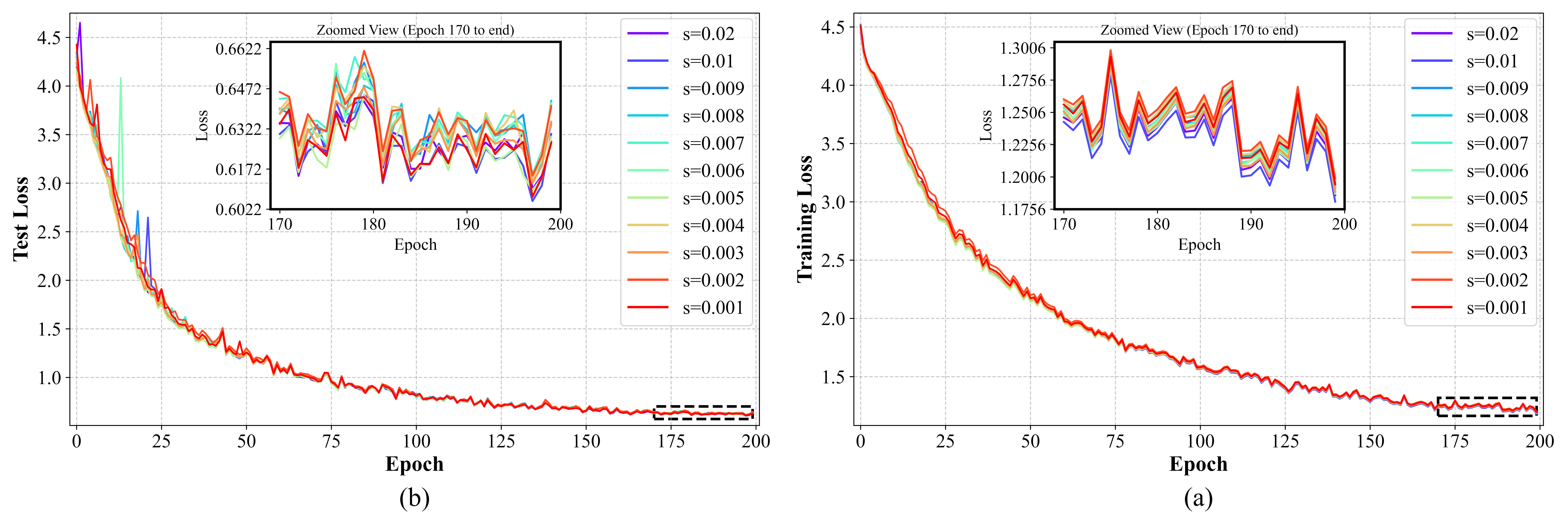}
\caption{EISAM hyperparameter sensitivity of $s$: (a) Training loss, (b) Test loss. In this set of experiments, we train a ResNet-50 model to classify the CIFAR-100 dataset. All models are trained with the same initial learning rate of 0.01, mini-batch size of 128, and weight decay of 1.0e-3 for 200 epochs. To explore the effect of hyperparameter $s$, we sweep s across a range of values: [0.001, 0.002, 0.003, 0.004, 0.005, 0.006, 0.007, 0.008, 0.009, 0.01, 0.02].}
\label{s2}
\end{figure*}

\subsection{Image Segmentation}

This section outlines the experimental setup designed to evaluate the EISAM optimizer for image segmentation tasks, benchmarked against five optimizers. A systematic hyperparameter search optimized key parameters with results averaged over three independent runs for robustness. 

\begin{table}[htbp]
    \centering
    \caption{Results on ISIC2018 using U-Net. Text marked as bold indicates the best result.}
    \label{SEG}
    \begin{tabular*}{\linewidth}{@{\extracolsep{\fill}} l c c c c}
        \toprule
        Optimizer & Validation & U-Net Score \\
        \midrule
        Adam & $88.46\pm0.21$ & $1.2698\pm0.01$ \\
        SAM & $88.47\pm0.38$ & $1.2594\pm0.05$ \\
        ASAM & $88.36\pm0.27$ & $1.2652\pm0.06$ \\
        GSAM & $87.58\pm0.56$ & $1.1989\pm0.06$ \\
        FSAM & $89.07\pm0.47$ & $\bf1.2973\pm0.03$ \\
        EISAM (Ours) & $\bf89.11\pm0.31$ & $1.2735\pm0.02$ \\
        \bottomrule
    \end{tabular*}
\end{table}

The experiment utilized the ISIC2018 medical image segmentation dataset and the U-Net network architecture. Performance was assessed using the best validation accuracy and U-Net Model Score metrics. The U-Net Model Score metric, employed in this study, is defined as the sum of the Jaccard Similarity Coefficient and the Dice Coefficient, providing a comprehensive measure of segmentation accuracy by combining overlap and similarity metrics. This composite metric effectively captures both the spatial overlap between predicted and ground-truth segmentations (via the Jaccard Similarity Coefficient) and the pixel-wise agreement (via the Dice Coefficient), offering a robust evaluation of segmentation quality. This experimental framework enables a fair and detailed comparison of EISAM against Adam, SAM, ASAM, GSAM and FSAM for image segmentation. The hyperparameter search and averaging over three runs ensure reliable and reproducible results. The experimental results are shown in Table \ref{SEG}. EISAM attained the highest validation accuracy across all optimizers on the ISIC2018 dataset. Nevertheless, its U-Net Score was marginally inferior to the top result achieved by FSAM.

\begin{figure*}[htbp]
\centering
\includegraphics[width=1.0\textwidth, angle=0, trim=0cm 0cm 0cm 0cm, clip]{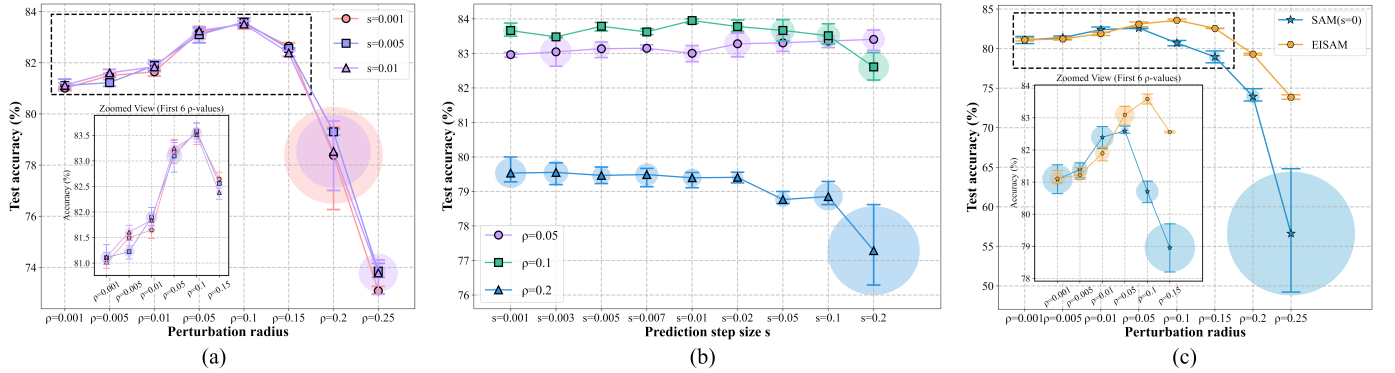}
\caption{Sensitivity analysis for EISAM and comparison with SAM: (a) impact of perturbation radius $\rho$ in EISAM, (b) impact of prediction step size $s$ in EISAM, and (c) comparison of $\rho$ sensitivity between SAM and EISAM.}
\label{rs3}
\end{figure*}

\subsection{Hyperparameter Sensitivity}

In this section, we evaluate the sensitivity of the EISAM optimizer to its two additional hyperparameters, $\rho$ (perturbation radius) and $s$ (prediction step size). These hyperparameters balance sharpness-aware optimization and extragradient-inspired updates, affecting model generalization.

Experiments were conducted on the CIFAR-100 dataset using a ResNet-50 architecture. We systematically varied $\rho$ across the range [0.001, 0.25] and $s$ across [0.001, 0.02], while keeping other training parameters fixed. The model was trained for 200 epochs, and performance was assessed using test accuracy and loss, averaged over three runs to account for variability.

\textbf{Impact of $\rho$}: The perturbation radius $\rho$ governs the extent of sharpness-aware perturbations in EISAM. Fig. \ref{rho2}(a) shows the training loss over 200 epochs for various $\rho$ values. Smaller $\rho$ values (e.g., 0.001–0.05) yield lower training losses, with narrow shaded regions (indicating standard deviation) in the zoomed-in view (epochs 170–200). In contrast, larger $\rho$ values (e.g., 0.2 to 0.25) result in higher training losses and greater fluctuations, indicating less stable convergence. Test loss, shown in Fig. \ref{rho2}(b), follows a similar pattern.

\textbf{Impact of $s$}: Fig. \ref{s2} shows that EISAM's training and test losses remain closely clustered and stable across different $s$ values. Even when paired with other hyperparameters like $\rho$, the EISAM optimizer demonstrates robustness, with minimal performance variation. This evidence highlights EISAM’s reliability and reduced need for fine-tuning $s$, making it a practical choice for deep learning optimization.

\textbf{Joint Effect of $  \rho  $ and $  s  $}: The combined influence of $  \rho  $ and $  s  $ is clearly illustrated in Fig. \ref{rs3}, which visualizes the sensitivity of the loss landscape to these two hyperparameters. The key advantage of EISAM stems from its prediction step, which explicitly explores the local geometry of the loss landscape before the correction step. This exploration is the primary mechanism behind both the improved generalization performance and the significantly reduced sensitivity to the perturbation radius $  \rho  $. Notably, when $  s=0  $, the prediction step is effectively disabled, causing the method to degenerate to standard SAM and recovering its higher sensitivity to $  \rho  $. The elliptical contour plot in Fig. \ref{rs3} is centered at the origin, with the upper and lower bounds representing the maximum and minimum loss values, respectively, and the contour size indicating the variance.

\section{Conclusion}
\label{5conclusion}

In this paper, we introduced the EISAM optimizer, a novel optimizer that improves generalization and training efficiency using a two-step update strategy inspired by the extragradient method. Extensive experiments on image classification, Natural Language Processing, object detection, and image segmentation tasks, using datasets such as CIFAR-10, CIFAR-100, ImageNet-1K, COCO2017, ISIC2018, and BOOLQ, showed that EISAM consistently outperformed baseline optimizers, including SGD, SAM, and Adam. In particular, EISAM achieved higher test accuracy and lower training loss across various model architectures while also exhibiting reduced sensitivity to hyperparameter tuning, particularly with respect to the perturbation radius $\rho$ and the prediction step size $s$.

Our theoretical analysis further supports these empirical findings, showing that EISAM effectively guides model parameters toward flatter minima, which are associated with improved generalization. The excess risk analysis provided a robust theoretical foundation, confirming EISAM's ability to balance convergence speed and generalization ability. The hyperparameter sensitivity study demonstrated EISAM’s flexibility and ease of tuning across tasks and datasets.

In summary, EISAM represents an advance in deep learning optimization, offering a scalable and broadly applicable solution that outperforms existing methods in both hyperparameter tuning efficiency and generalization. Its ability to adapt to various architectures and tasks positions EISAM as a promising tool for researchers and practitioners in deep learning. 

\bibliographystyle{IEEEtran}
\bibliography{main}

@article{Hochreiter1997,
  author = {Hochreiter, Sepp and Schmidhuber, J{\"u}rgen},
  title = {Flat minima},
  journal = {Neural Computation},
  volume = {9},
  number = {1},
  year = {1997},
  pages = {1--42}
}

@inproceedings{Dinh2017,
  author    = {Laurent Dinh and Razvan Pascanu and Samy Bengio and Yoshua Bengio},
  title     = {Sharp minima can generalize for deep nets},
  booktitle = {Proc. Int. Conf. Mach. Learn.},
  pages     = {1019--1028},
  year      = {2017}
}

@inproceedings{Keskar2017,
  author    = {Nitish Shirish Keskar and Dheevatsa Mudigere and Jorge Nocedal and Mikhail Smelyanskiy and Ping Tak Peter Tang},
  title     = {On large-batch training for deep learning: Generalization gap and sharp minima},
  booktitle = {Proc. Int. Conf. Learn. Represent.},
  year      = {2017}
}

@inproceedings{Foret2021,
  author    = {Pierre Foret and Ariel Kleiner and Hossein Mobahi and Behnam Neyshabur},
  title     = {Sharpness-aware minimization for efficiently improving generalization},
  booktitle = {Proc. Int. Conf. Learn. Represent.},
  year      = {2021}
}

@article{Nguyen2018,
  author  = {Trong Phong Nguyen and Edouard Pauwels and {\'E}mile Richard and Bruce W. Suter},
  title   = {Extragradient method in optimization: Convergence and complexity},
  journal = {J. Optim. Theory Appl.},
  volume  = {176},
  number  = {1},
  pages   = {137--162},
  year    = {2018}
}

@techreport{krizhevsky2009cifar,
  author      = {Alex Krizhevsky},
  title       = {Learning multiple layers of features from tiny images},
  institution = {University of Toronto},
  number      = {Tech. Rep. TR-2009},
  year        = {2009},
  pages       = {1--60}
}

@article{deng2009imagenet,
  author  = {Jia Deng and Wei Dong and Richard Socher and Li-Jia Li and Kai Li and Li Fei-Fei},
  title   = {{ImageNet}: a large-scale hierarchical image database},
  journal = {Proc. IEEE},
  volume  = {97},
  number  = {11},
  pages   = {248--255},
  year    = {2009}
}

@inproceedings{lin2014coco,
  author    = {Tsung-Yi Lin and Michael Maire and Serge Belongie and James Hays and Pietro Perona and Deva Ramanan and Piotr Doll{\'a}r and C. Lawrence Zitnick},
  title     = {{Microsoft COCO}: common objects in context},
  booktitle = {Computer Vision -- ECCV 2014},
  series    = {Lecture Notes in Computer Science},
  volume    = {8693},
  pages     = {740--755},
  year      = {2014},
}

@misc{codella2019isic,
  author  = {Noel Codella and Veronica Rotemberg and Philipp Tschandl and M. Emre Celebi and Stephen Dusza and David Gutman and Brian Helba and Aadi Kalloo and Konstantinos Liopyris and Michael Marchetti and Harald Kittler and Allan Halpern},
  title   = {Skin lesion analysis toward melanoma detection 2018: A challenge hosted by the International Skin Imaging Collaboration ({ISIC})},
  url ={https://arxiv.org/abs/1902.03368},
  year    = {2019}
}

@article{clark2019boolq,
  author  = {Christopher Clark and Kenton Lee and Ming-Wei Chang and Tom Kwiatkowski and Michael Collins and Kristina Toutanova},
  title   = {{BoolQ}: exploring the surprising difficulty of natural yes/no questions},
  journal = {Trans. Assoc. Comput. Linguistics},
  volume  = {7},
  pages   = {2924--2943},
  year    = {2019}
}

@inproceedings{Andriushchenko2022,
  author    = {Maksym Andriushchenko and Nicolas Flammarion},
  title     = {Towards understanding sharpness-aware minimization},
  booktitle = {Proc. Int. Conf. Mach. Learn.},
  series    = {Proc. Mach. Learn. Res.},
  volume    = {162},
  pages     = {639--668},
  year      = {2022}
}

@inproceedings{Chaudhari2017,
  author  = {Pratik Chaudhari and Anna Choromanska and Stefano Soatto and Yann LeCun and Carlo Baldassi and Christian Borgs and Jennifer Chayes and Levent Sagun and Riccardo Zecchina},
  title   = {Entropy-{SGD}: Biasing Gradient Descent into Wide Valleys},
  booktitle = {Proc. Int. Conf. Learn. Represent. (ICLR)},
  year    = {2017}
}

@inproceedings{Li2018,
  author = {Li, Hao and Xu, Zheng and Taylor, Gavin and Studer, Christoph and Goldstein, Tom},
  title = {Visualizing the loss landscape of neural nets},
  booktitle = {Proc. Adv. Neural Inf. Process. Syst.},
  volume    = {31},
  pages     = {6391--6401},
  year      = {2018}
}

@inproceedings{Kwon2021,
  author = {Kwon, Jungmin and Kim, Jeongseop and Park, Hyunnseo and Choi, In-Kwon},
  title = {{ASAM}: Adaptive sharpness-aware minimization for scale-invariant learning of deep neural networks},
  booktitle = {Proc. Int. Conf. Mach. Learn.},
  series    = {Proc. Mach. Learn. Res.},
  volume    = {139},
  pages     = {5905--5914},
  year      = {2021}
}

@inproceedings{Du2020,
  author = {Du, Jiawei and Zhang, Hu and Zhou, Joet Tianyi and Yang, Yi and Feng, Jiashi},
  title = {Query-efficient meta attack to deep neural networks},
  booktitle = {Proc. Int. Conf. Learn. Represent. (ICLR)},
  year = {2020}
}

@article{Nesterov1983,
  author  = {Nesterov, Yurii},
  title   = {A Method for Solving the Convex Programming Problem with Convergence Rate {$O(1/k^2)$}},
  journal = {Dokl. Akad. Nauk SSSR},
  volume  = {269},
  number  = {3},
  year    = {1983},
  pages   = {543--547}
}

@article{Bolte2017,
  author = {Bolte, J{\'e}r{\^o}me and Nguyen, Trong Phong and Peypouquet, Juan and Suter, Bruce W.},
  title = {From error bounds to the complexity of first-order descent methods for convex functions},
  journal = {Math. Program.},
  volume = {165},
  number = {2},
  year = {2017},
  pages = {471--507}
}

@inproceedings{McAllester1999,
  author    = {McAllester, David A.},
  title     = {{PAC}-Bayesian model averaging},
  booktitle = {Proc. 12th Annu. Conf. Comput. Learn. Theory},
  pages     = {164--170},
  publisher = {ACM Press},
  year      = {1999}
}

@inproceedings{jiang2019,
  author    = {Jiang, Yiding and Neyshabur, Behnam and Mobahi, Hossein and Krishnan, Dilip and Bengio, Samy},
  title     = {Fantastic Generalization Measures and Where to Find Them},
  booktitle = {Proc. Int. Conf. Learn. Represent. (ICLR)},
  year      = {2020},
}

@inproceedings{He2016,
  author    = {He, Kaiming and Zhang, Xiangyu and Ren, Shaoqing and Sun, Jian},
  title     = {Deep Residual Learning for Image Recognition},
  booktitle = {Proc. IEEE Conf. Comput. Vis. Pattern Recognit. (CVPR)},
  year      = {2016},
  pages     = {770--778},
}

@inproceedings{Han2017,
  author    = {Han, Dongyoon and Kim, Jiwhan and Kim, Junmo},
  title     = {Deep Pyramidal Residual Networks},
  booktitle = {Proc. IEEE Conf. Comput. Vis. Pattern Recognit. (CVPR)},
  year      = {2017},
  pages     = {6307-6315},
}

@inproceedings{Zagoruyko2016,
  author = {Zagoruyko, Sergey and Komodakis, Nikos},
  title = {Wide residual networks},
  booktitle = {Proc. Brit. Mach. Vis. Conf.},
  year = {2016}
}

@inproceedings{Cubuk2019,
  author = {Cubuk, Ekin D. and Zoph, Barret and Mane, Dandelion and Vasudevan, Vijay and Le, Quoc V.},
  title = {{AutoAugment}: Learning augmentation strategies from data},
  booktitle = {Proc. IEEE/CVF Conf. Comput. Vis. Pattern Recognit. (CVPR)},
  year      = {2019},
  pages     = {113--123},
}

@inproceedings{Yun2019,
  author = {Yun, Sangdoo and Han, Dongyoon and Oh, Seong Joon and Chun, Sanghyuk and Choe, Junsuk and Yoo, Youngjoon},
  title = {{CutMix}: Regularization strategy to train strong classifiers with localizable features},
  booktitle = {Proc. IEEE/CVF Int. Conf. Comput. Vis. (ICCV)},
  year      = {2019},
  pages     = {6023--6032},
}

@inproceedings{Muller2019,
  author = {M{\"u}ller, Rafael and Kornblith, Simon and Hinton, Geoffrey E.},
  title = {When does label smoothing help?},
  booktitle = {Proc. Adv. Neural Inf. Process. Syst.},
  year = {2019},
}

@inproceedings{Zhang2018,
  author = {Zhang, Hongyi and Cisse, Moustapha and Dauphin, Yann N. and Lopez-Paz, David},
  title = {mixup: Beyond empirical risk minimization},
  booktitle = {Proc. Int. Conf. Learn. Represent. (ICLR)},
  year = {2018}
}

@inproceedings{Kingma2015,
  author    = {Kingma, Diederik P. and Ba, Jimmy},
  title     = {Adam: A Method for Stochastic Optimization},
  booktitle = {Proc. Int. Conf. Learn. Represent. (ICLR)},
  year      = {2015}
}

@inproceedings{Loshchilov2017,
  author = {Loshchilov, Ilya and Hutter, Frank},
  title = {{SGDR}: Stochastic gradient descent with warm restarts},
  booktitle = {Proc. Int. Conf. Learn. Represent. (ICLR)},
  year = {2017}
}

@inproceedings{Zhang2019,
  author = {Zhang, Michael R. and Lucas, James and Ba, Jimmy and Hinton, Geoffrey},
  title = {Lookahead optimizer: k steps forward, 1 step back},
  booktitle = {Proc. Adv. Neural Inf. Process. Syst. (NeurIPS)},
  year      = {2019},
}

@article{Scabini2023,
  author  = {Scabini, Leonardo and Zielinski, Kallil M. and Ribas, Lucas C. and Gon{\c c}alves, Wesley N. and De Baets, Bernard and Bruno, Odemir M.},
  title   = {{RADAM}: Texture Recognition Through Randomized Aggregated Encoding of Deep Activation Maps},
  journal = {Pattern Recognit.},
  volume  = {143},
  year    = {2023},
  pages   = {109802}
}

@article{Korpelevich1976,
  author  = {Korpelevich, G. M.},
  title   = {The Extragradient Method for Finding Saddle Points and Other Problems},
  journal = {Math. Notes (English Transl. of Matematicheskii Sbornik)},
  volume  = {12},
  year    = {1976},
  pages   = {747--756}
}

@article{Luo1993,
  author  = {Luo, Zhi-Quan and Tseng, Paul},
  title   = {Error Bounds and Convergence Analysis of Feasible Descent Methods: A General Approach},
  journal = {Ann. Oper. Res.},
  volume  = {46},
  number  = {1},
  pages   = {157--178},
  year    = {1993}
}

@article{Attouch2009,
  author  = {Attouch, Hedy and Bolte, J{\'e}r{\^o}me},
  title   = {On the Convergence of the Proximal Algorithm for Nonsmooth Functions Involving Analytic Features},
  journal = {Math. Program.},
  volume  = {116},
  number  = {1},
  pages   = {5--16},
  year    = {2009}
}

@article{Attouch2013,
  author  = {Attouch, Hedy and Bolte, J{\'e}r{\^o}me and Svaiter, Benar F.},
  title   = {Convergence of Descent Methods for Semi-Algebraic and Tame Problems: Proximal Algorithms, Forward--Backward Splitting, and Regularized Gauss--Seidel Methods},
  journal = {Math. Program.},
  volume  = {137},
  number  = {1-2},
  pages   = {91--129},
  year    = {2013}
}

@article{Bolte2014,
  author  = {Bolte, J{\'e}r{\^o}me and Sabach, Shoham and Teboulle, Marc},
  title   = {Proximal Alternating Linearized Minimization for Nonconvex and Nonsmooth Problems},
  journal = {Math. Program.},
  volume  = {146},
  number  = {1-2},
  pages   = {459--494},
  year    = {2014}
}

@article{Beck2009,
  author = {Beck, Amir and Teboulle, Marc},
  title   = {A Fast Iterative Shrinkage-Thresholding Algorithm for Linear Inverse Problems},
  journal = {SIAM J. Imaging Sci.},
  volume  = {2},
  number  = {1},
  pages   = {183--202},
  year    = {2009}
}

@inproceedings{ghorbani2019,
  author    = {Ghorbani, Behrooz and Krishnan, Shankar and Xiao, Ying},
  title     = {An Investigation into Neural Net Optimization via Hessian Eigenvalue Density},
  booktitle = {Proc. Int. Conf. Mach. Learn. (ICML)},
  year      = {2019},
  volume    = {97},
  pages     = {2232--2241}
}

@article{Bousquet2002,
  author = {Bousquet, Olivier and Elisseeff, Andr{\'e}},
  title = {Stability and generalization},
  journal = {J. Mach. Learn. Res.},
  volume = {2},
  year = {2002},
  pages = {499--526}
}

@inproceedings{Hardt2016,
  author = {Hardt, Moritz and Recht, Benjamin and Singer, Yoram},
  title = {Train faster, generalize better: Stability of stochastic gradient descent},
  booktitle = {Proc. Int. Conf. Mach. Learn. (ICML)},
  year = {2016},
}

@inproceedings{ren2017fasterrcnn,
  author = {Ren, Shaoqing and He, Kaiming and Girshick, Ross and Sun, Jian},
  title = {{Faster R-CNN}: Towards real-time object detection with region proposal networks},
  booktitle = {Proc. Adv. Neural Inf. Process. Syst. (NeurIPS)},
  volume = {28},
  year = {2015}
}

@inproceedings{dosovitskiy2021vit,
  author    = {A. Dosovitskiy and L. Beyer and A. Kolesnikov and D. Weissenborn and X. Zhai and T. Unterthiner and M. Dehghani and M. Minderer and G. Heigold and S. Gelly and J. Uszkoreit and N. Houlsby},
  title     = {An Image is Worth 16x16 Words: Transformers for Image Recognition at Scale},
  booktitle = {Proc. Int. Conf. Learn. Represent. (ICLR)},
  year      = {2021}
}

@inproceedings{ronneberger2015unet,
  author = {Ronneberger, Olaf and Fischer, Philipp and Brox, Thomas},
  title     = {{U-Net}: Convolutional Networks for Biomedical Image Segmentation},
  booktitle = {Int. Conf. Med. Image Comput. Comput.-Assist. Interv. (MICCAI)},
  year      = {2015},
  pages     = {234--241}
}

@article{raffel2020t5,
  author = {Raffel, Colin and Shazeer, Noam and Roberts, Adam and Lee, Katherine and Narang, Sharan and Matena, Michael and Zhou, Yanqi and Li, Wei and Liu, Peter J.},
  title = {Exploring the limits of transfer learning with a unified text-to-text transformer},
  journal = {J. Mach. Learn. Res.},
  volume = {21},
  number = {140},
  year = {2020},
  pages = {1--67}
}

@article{Tan2024,
  author = {Tan, Chengli and Zhang, Jiangshe and Liu, Junmin and Gong, Yihong},
  title = {Sharpness-aware lookahead for accelerating convergence and improving generalization},
  journal = {IEEE Trans. Pattern Anal. Mach. Intell.},
  volume = {46},
  number = {12},
  year = {2024},
  pages = {10375--10388}
}

@inproceedings{du2024friendly,
   author    = {Tao Li and Pan Zhou and Zhengbao He and Xinwen Cheng and Xiaolin Huang},
  title     = {Friendly Sharpness-Aware Minimization},
  booktitle = {Proc. IEEE/CVF Conf. Comput. Vis. Pattern Recognit. (CVPR)},
  year      = {2024}
}

@inproceedings{zhuang2022surrogate,
  author = {Juntang Zhuang and Boqing Gong and Liangzhe Yuan and Yin Cui and Hartwig Adam and Nicha Dvornek and Sekhar Tatikonda and James Duncan, Ting Liu},
  title = {Surrogate Gap Minimization Improves Sharpness-Aware Training},
  booktitle = {Proc. Int. Conf. Learn. Represent. (ICLR)},
  year = {2022}
}

@inproceedings{gupta2019lvis,
  title={{LVIS}: A Dataset for Large Vocabulary Instance Segmentation},
  author={Gupta, Agrim and Doll{\'a}r, Piotr and Girshick, Ross},
  booktitle={Proc. IEEE/CVF Conf. Comput. Vis. Pattern Recognit. (CVPR)},
  year={2019}
}

@inproceedings{oikonomou2025sharpness,
  title={Sharpness-Aware Minimization: General Analysis and Improved Rates},
  author={Dimitris Oikonomou and Nicolas Loizou},
  booktitle={Proc. Int. Conf. Learn. Represent. (ICLR)},
  year={2025}
}

@inproceedings{zhou2025extragradient,
  title={Extragradient Preference Optimization ({EGPO}): Beyond Last-Iterate Convergence for Nash Learning from Human Feedback},
  author={Runlong Zhou and Maryam Fazel and Simon S. Du},
  booktitle={Proc. Conf. Lang. Model. (COLM)},
  year={2025}
}

@article{hassan2025sharpness,
  title={Do Sharpness-Based Optimizers Improve Generalization in Medical Image Analysis?},
  author={Mohamed Hassan and Aleksandar Vakanski and Boyu Zhang and Min Xian},
  journal={IEEE Access},
  year={2025},
  volume={13},
  pages={82972-82985}
}

@inproceedings{andriushchenko2023modern,
  title     = {A Modern Look at the Relationship between Sharpness and Generalization},
  author    = {Andriushchenko, Maksym and Croce, Francesco and M{\"u}ller, Maximilian and Hein, Matthias and Flammarion, Nicolas},
  booktitle = {Proc. Int. Conf. Mach. Learn. (ICML)},
  year      = {2023},
}

@misc{schliserman2025flat,
  title     = {Flat Minima and Generalization: Insights from Stochastic Convex Optimization},
  author    = {Schliserman, Matan and Vansover-Hager, Shira and Koren, Tomer},
  year      = {2025},
  url ={https://arxiv.org/abs/2511.03548}
}

@misc{lell2023split,
  title     = {The Split Matters: Flat Minima Methods for Improving the Performance of GNNs},
  author    = {Lell, Nicolas and Scherp, Ansgar},
  year      = {2023},
  url ={https://arxiv.org/abs/2306.09121}
}

@article{Rosenbrock1960AnAM,
  title={An Automatic Method for Finding the Greatest or Least Value of a Function},
  author={Howard Rosenbrock},
  journal={Comput. J.},
  year={1960},
  volume={3},
  pages={175-184}
}

\begin{IEEEbiography}[{\includegraphics[width=1.0in,height=1.15in,clip,keepaspectratio]{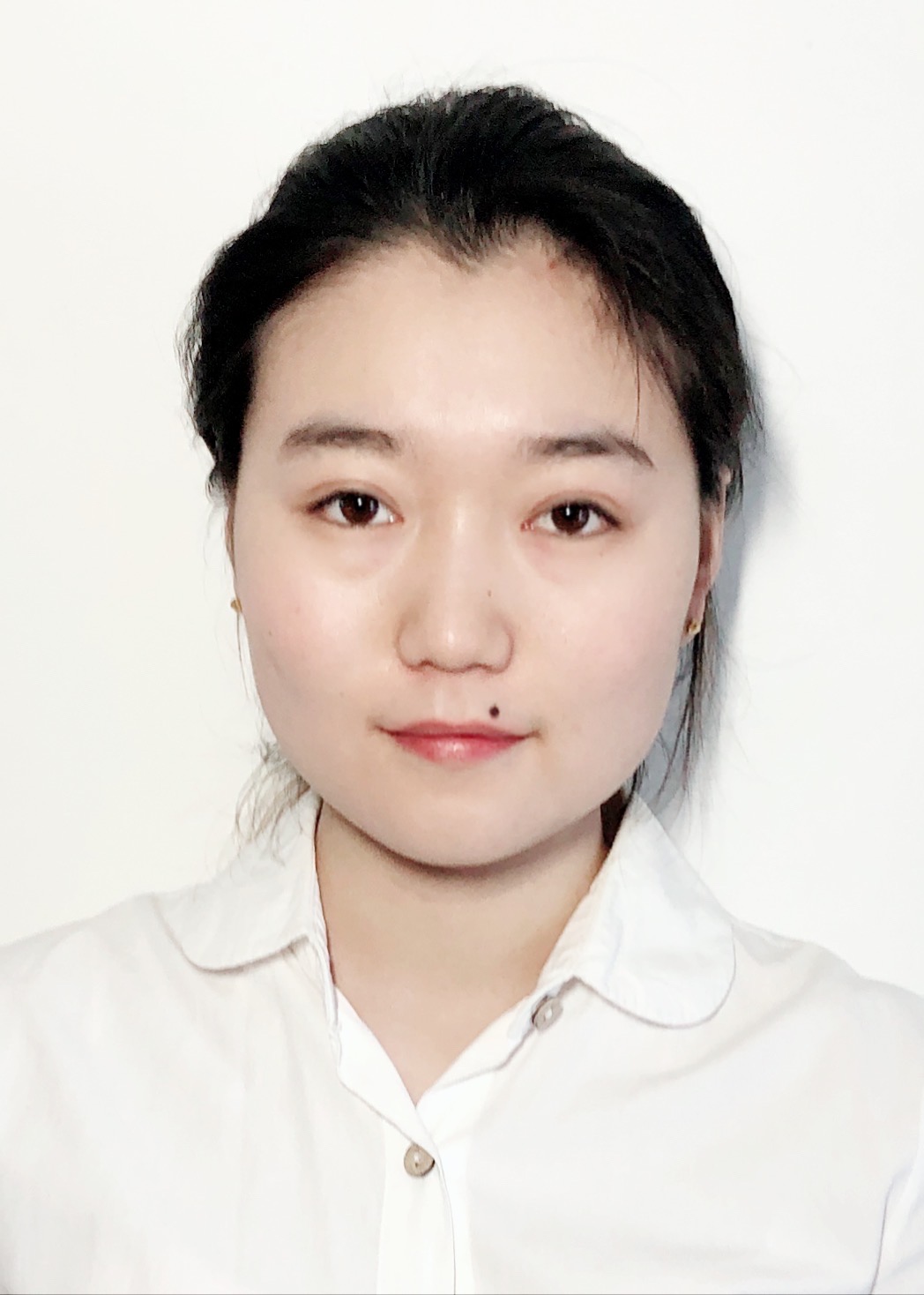}}]{Yao~Fu} received the Master degree in Probability and Statistics from Northeastern University, Shenyang, China, in 2017. She is currently pursuing a Ph.D. degree in Statistics at Xi’an Jiaotong University, Xi’an, China. Her current research interests include optimization, deep learning, and related areas.
\end{IEEEbiography}
\vspace{-1.1cm}
\begin{IEEEbiography}[{\includegraphics[width=1.0in,height=1.15in,clip,keepaspectratio]{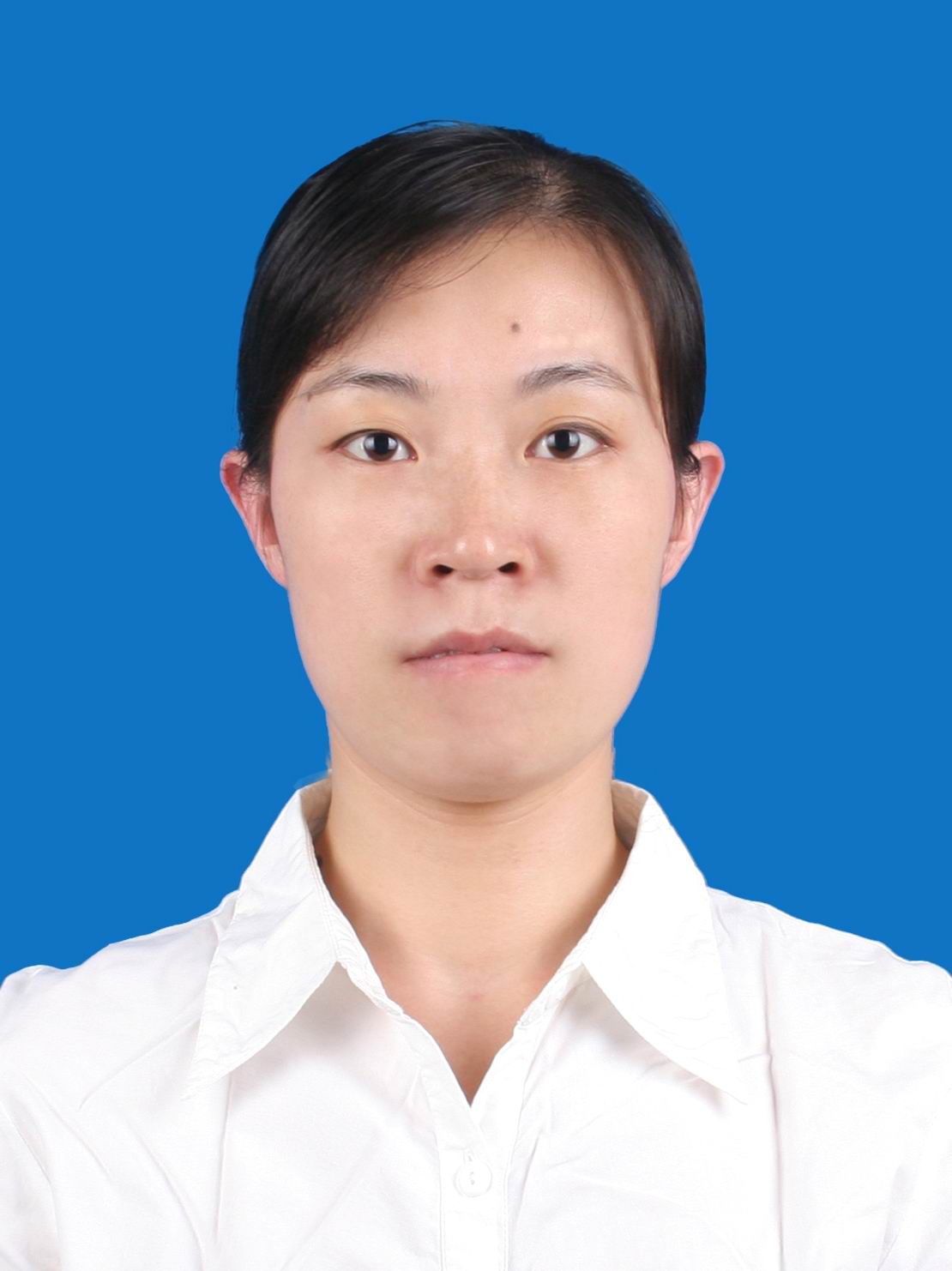}}]{Chunxia~Zhang} received her Ph.D degree in Applied Mathematics from Xi’an Jiaotong University, Xi’an, China, in 2010. At present, she is a Professor with the School of Mathematics and Statistics at Xi’an Jiaotong University. She has authored and coauthored about 50 journal papers on ensemble learning, high-dimensional data analysis, deep learning and etc. Her main interests include deep learning, Bayesian statistics, seismic data processing and etc.
\end{IEEEbiography}
\vspace{-1.1cm}
\begin{IEEEbiography}[{\includegraphics[width=1.0in,height=1.15in,clip,keepaspectratio]{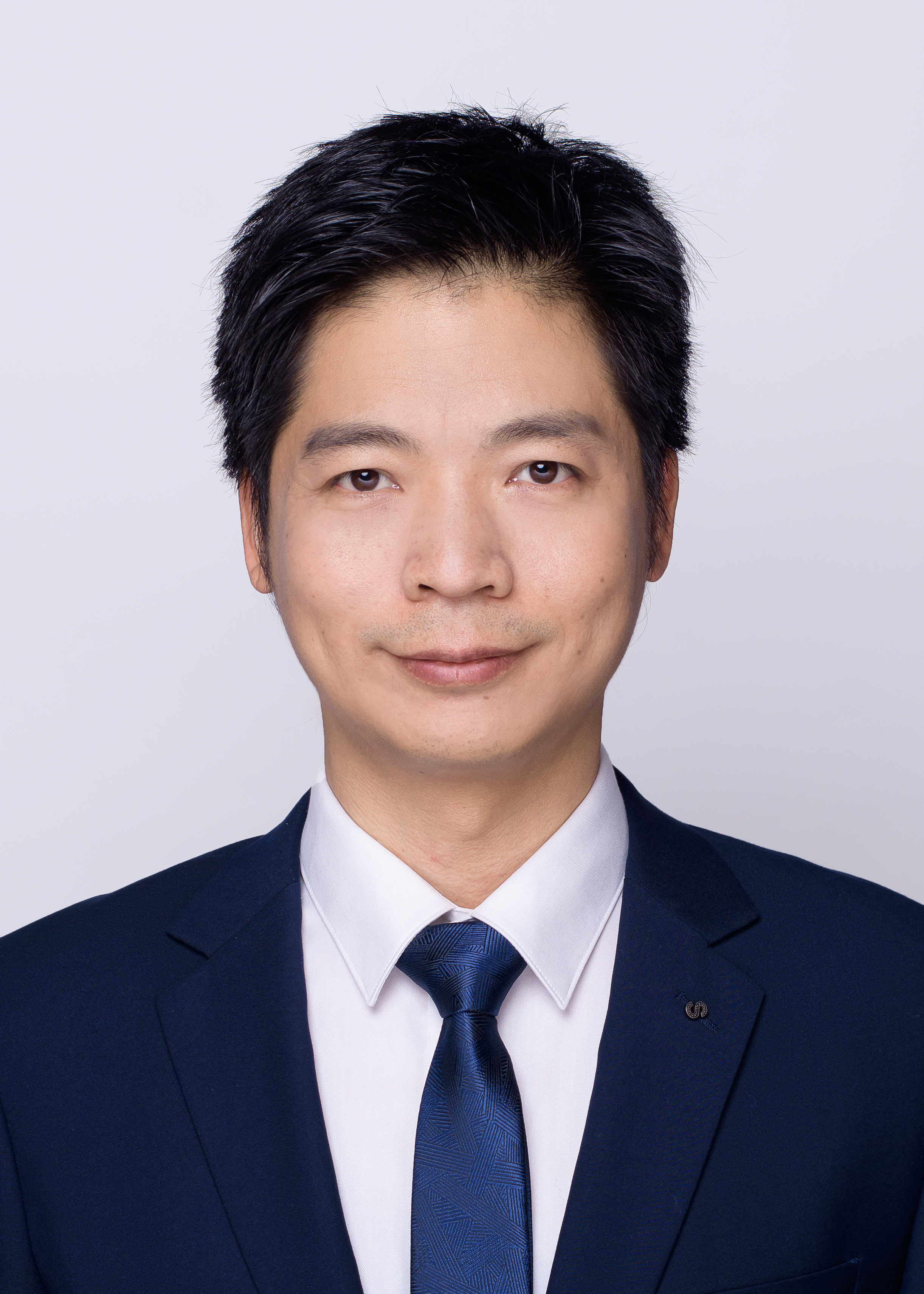}}]{Junmin~Liu} received the Ph.D degree in Mathematics from Xi’an Jiaotong University, Xi’an, China, in 2013. He is currently a Professor with the School of Mathematics and Statistics, Xi’an Jiaotong University, Xi’an, China.
His research interests primarily focus on machine learning, deep learning, and computer vision, with applications in remotely sensed image fusion, hyperspectral unmixing, and object detection. He has authored over 60 research papers published in prestigious international conferences and journals. Dr. Liu is an active member of IEEE and the Chinese Society for Industrial and Applied Mathematics (CSIAM). He serves as a Topic Editor for Remote Sensing.
\end{IEEEbiography}
\vspace{-1.1cm}
\begin{IEEEbiography}[{\includegraphics[width=1.0in,height=1.15in,clip,keepaspectratio]{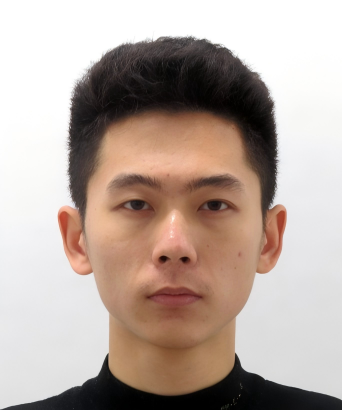}}]{Yihang~Jin} obtained a Bachelor degree from Nanjing University of Information Science and Technology in 2018. He is currently pursuing a Master degree at Xi 'an Jiaotong University. His current research interests include related fields such as association rule mining, large language models and optimization.
\end{IEEEbiography}
\vspace{-1.1cm}
\begin{IEEEbiography}[{\includegraphics[width=1.0in,height=1.15in,clip,keepaspectratio]{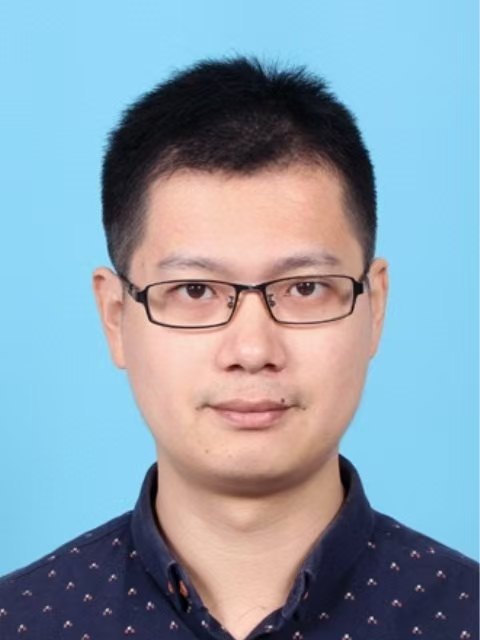}}]{Haishan~Ye} received the Ph.D degree from the Department of
Computer Science of Shanghai Jiao Tong University in 2018. He is currently an Associate Professor with the School of Management, Xi’an Jiaotong University. His research interests include artificial intelligence, machine learning, and data mining. He is especially interested in statistical methods, randomized linear algebra, optimization problems and deep learning.
\end{IEEEbiography}
\vspace{-1.1cm}
\begin{IEEEbiography}
[{\includegraphics[width=1.0in,height=1.15in,clip,keepaspectratio]{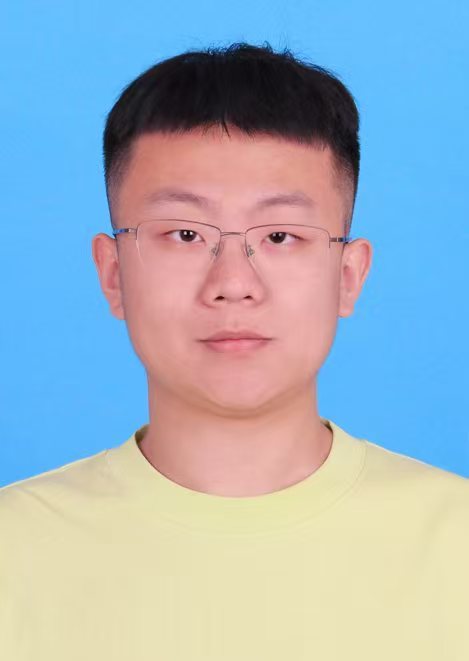}}]{Yuanao~Yang} is currently pursuing a Master degree of State Key Laboratory of Multiphase Flow in Power Engineering, Xi'an Jiaotong University. He is mainly engaged in the research on the Combination of Machine Learning and Flow Heat Transfer.
\end{IEEEbiography}
\clearpage

\appendices
\setcounter{page}{1}

\section{Additional Material for Section \ref{sec1}} 
\label{appa}

\begin{figure*}[h]
    \centering 
    \includegraphics[width=1\textwidth, trim=0.0cm 0.0cm 0.0cm 0.0cm, 
    clip,angle=0]{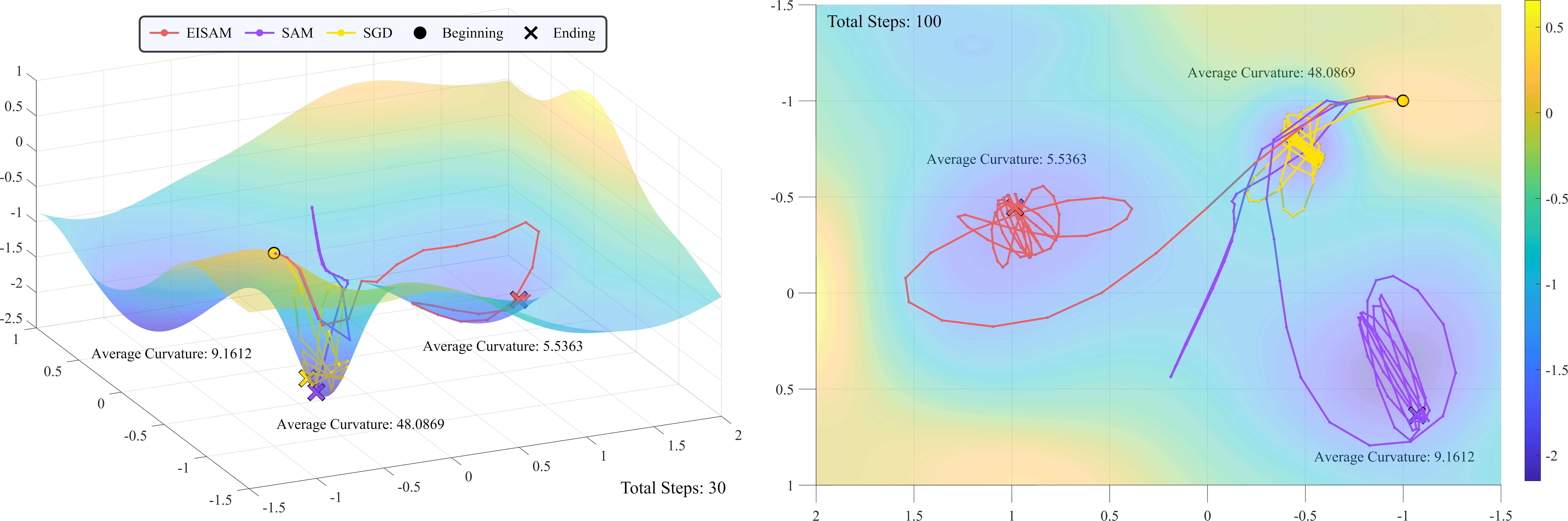}	
    \caption{Visualization of training trajectories in the loss landscape. Lower average curvature indicates flatter minima and better expected generalization. EISAM achieves lower curvature than SAM and SGD across different runs.} 
    \label{3d-appa}
\end{figure*}

To provide deeper insights into the behavior of different optimizers on nonconvex loss landscapes, we visualize the training trajectories of EISAM, SAM, and SGD in Fig. \ref{3d-appa}. The figure presents two views: a 3D visualization over 30 steps and a projection onto a 2D subspace of the parameter space over 100 steps. In the 100-step visualization, the background color map indicates local curvature—warmer colors denote higher curvature (sharper regions), while cooler colors represent lower curvature (flatter regions).

In the 30-step visualization (left panel), SGD (yellow) rapidly converges but quickly becomes trapped in a sharp local minimum with high average curvature, reflecting its tendency to settle in narrow, high-curvature regions. SAM (purple) explores more broadly but has not yet reached a favorable flat region. In contrast, EISAM (red) efficiently identifies a direction toward flatter regions early in training, advancing significantly toward the lowest-curvature basin with smoother progress, demonstrating faster initial convergence toward promising flat minima.

In the 100-step visualization (right panel), the advantages of EISAM become even more pronounced. SGD remains stuck in the same sharp minimum (average curvature 48.0869), unable to escape due to the steep barriers surrounding it, which explains its poorer generalization performance. SAM reaches a flatter region (average curvature 9.1612) compared to SGD but exhibits larger oscillations and follows more rigid, linear-like segments in its trajectory. This patterned behavior indicates that once SAM encounters complex landscape features (e.g., saddle points or elongated valleys), it undergoes repetitive escapes, leading to instability and suboptimal exploration. EISAM, however, directly navigates to the flattest region (average curvature 5.5363), with a more stable and direct path, avoiding excessive fluctuations and demonstrating better robustness in complex, rugged loss landscapes.

These trajectories highlight EISAM’s key strengths: rapid identification of flat directions, stable progression, and effective escape from sharp or complex regions, ultimately leading to flatter minima associated with better generalization.

\section{Additional Material for Section \ref{sec4}}
\label{appb}
In this section, we provide detailed proofs for the lemmas and theorems. All proofs rely on the assumptions defined in Section 4 (Strong Convexity, Smoothness, and Lipschitz Continuity). For clarity, we recall key notations: $\mathbf{w}_t$ and $\mathbf{v}_t$ are parameter trajectories on datasets $S$ and $S'$; $\delta_t = \|\mathbf{w}_t - \mathbf{v}_t\|$; $\mathbf{H}_t = \nabla^2 F_{B_t}(\mathbf{w}_t)$ is the Hessian approximation; $\mu, L, G > 0$ are constants from the definitions.

\subsection{Proof of Lemma \ref{lem:identical} (Parameter Trajectory Difference for Identical Mini-Batches)}
With probability $  1 - \frac{1}{n}  $, the mini-batches $  B_t  $ are identical for both trajectories. The update rules are:
\begin{align*}
\mathbf{w}_{t+1} = \mathbf{y}_t - \eta \nabla F_{B_t}(\mathbf{w}_t'), \mathbf{v}_{t+1} = \mathbf{y}_t' - \eta \nabla F_{B_t}(\mathbf{v}_t'),
\end{align*}
where
\begin{align*}
\mathbf{y}_t = \mathbf{w}_t - s \nabla F_{B_t}(\mathbf{w}_t), \mathbf{w}_t' = \mathbf{y}_t + \rho \frac{\nabla F_{B_t}(\mathbf{w}_t)}{\|\nabla F_{B_t}(\mathbf{w}_t)\|}, 
\end{align*}
and similarly for $ \mathbf{y}_t' $ and $ \mathbf{v}_t' $. After computing $\mathbf{y}_t$, we update $\mathbf{w}_t \leftarrow \mathbf{y}_t$, and the perturbation uses the initial gradient at the original $\mathbf{w}_t$. 

Assume the Hessian matrix $ \mathbf{H}_t = \nabla^2 F_{B_t} $ is Lipschitz continuous with constant $ K $, i.e., $ \|\nabla^2 F_{B_t}(\mathbf{w}) - \nabla^2 F_{B_t}(\mathbf{v})\| \leq K \|\mathbf{w} - \mathbf{v}\| $ for all $ \mathbf{w}, \mathbf{v} $. Additionally, higher-order terms are bounded such that the Taylor remainder satisfies $ \|R_t\| \leq \frac{K}{2} \delta_t^2 $, where $ \delta_t = \|\mathbf{w}_t - \mathbf{v}_t\| $.
First, we need to bound the intermediate point difference $ \|\mathbf{y}_t - \mathbf{y}_t'\| $. Since:
\begin{align*}
\mathbf{y}_t - \mathbf{y}_t' = (\mathbf{w}_t - \mathbf{v}_t) - s (\nabla F_{B_t}(\mathbf{w}_t) - \nabla F_{B_t}(\mathbf{v}_t)).
\end{align*}
Using a first-order Taylor expansion around $ \mathbf{w}_t $ with remainder,
\begin{align*}
\nabla F_{B_t}(\mathbf{v}_t) = \nabla F_{B_t}(\mathbf{w}_t) + \mathbf{H}_t (\mathbf{v}_t - \mathbf{w}_t) + R_t,
\end{align*}
where $ \|R_t\| \leq \frac{K}{2} \delta_t^2 $. Thus,
\begin{align*}
\mathbf{y}_t - \mathbf{y}_t' &= (\mathbf{w}_t - \mathbf{v}_t) - s [\mathbf{H}_t (\mathbf{w}_t - \mathbf{v}_t) - R_t] \\
&= (\mathbf I - s \mathbf{H}_t) (\mathbf{w}_t - \mathbf{v}_t) + s R_t,
\end{align*}
where $ \mathbf I $ is the identity matrix. Taking norms,
\begin{align*}
\|\mathbf{y}_t - \mathbf{y}_t'\| \leq \|\mathbf I - s \mathbf{H}_t\| \cdot \delta_t + s \|R_t\| \leq (1 - s \mu) \delta_t + s \cdot \frac{K}{2} \delta_t^2,
\end{align*}
by strong convexity and smoothness (eigenvalues of $ \mathbf{H}_t $ satisfy $ \mu \leq \sigma_{\min}(\mathbf{H}_t) \leq \sigma_{\max}(\mathbf{H}_t) \leq L $), assuming $ s < 1/L $.
Next, for the perturbed points,
\begin{align*}
\mathbf{w}_t' - \mathbf{v}_t' = (\mathbf{y}_t - \mathbf{y}_t') + \rho \left( \frac{\nabla F_{B_t}(\mathbf{w}_t)}{\|\nabla F_{B_t}(\mathbf{w}_t)\|} - \frac{\nabla F_{B_t}(\mathbf{v}_t)}{\|\nabla F_{B_t}(\mathbf{v}_t)\|} \right). 
\end{align*}
The difference in unit vectors is bounded by the Lipschitz constant of the normalization function, which is $ O(1/G) $ assuming bounded gradients $ \|\nabla F_{B_t}\| \geq G > 0 $. Thus, the second term is bounded by $ O(\rho (1 - s \mu) \delta_t / G + \rho s K \delta_t^2 / (2G)) $. Overall,
\begin{align*}
\|\mathbf{w}_t' - \mathbf{v}_t'\| \leq (1 - s \mu) \delta_t + O( sK/2 \delta_t^2 + \rho \delta_t / G + \rho s K \delta_t^2 / G).
\end{align*}
Now, apply the co-coercivity lemma from strong convexity and smoothness \cite{Hardt2016}:
\begin{align*}
&\langle \nabla F_{B_t}(\mathbf{w}_t') - \nabla F_{B_t}(\mathbf{v}_t'), \mathbf{w}_t' - \mathbf{v}_t' \rangle \geq \frac{\mu L}{\mu + L} \|\mathbf{w}_t' - \mathbf{v}_t'\|^2 \\
&+ \frac{1}{\mu + L} \|\nabla F_{B_t}(\mathbf{w}_t') - \nabla F_{B_t}(\mathbf{v}_t')\|^2.
\end{align*}
Focusing on the first term,
\begin{equation*}
\langle \nabla F_{B_t}(\mathbf{w}_t') - \nabla F_{B_t}(\mathbf{v}_t'), \mathbf{w}_t' - \mathbf{v}_t' \rangle \geq \frac{\mu L}{\mu + L} [(1 - s \mu) \delta_t + O(\delta_t^2)]^2,
\end{equation*}
which expands to $ \frac{\mu L}{\mu + L} (1 - s \mu)^2 \delta_t^2 + O(\delta_t^3) $.
For the squared distance update,
\begin{align*}
\delta_{t+1}^2 &= \|\mathbf{y}_t - \mathbf{y}_t'\|^2 - 2\eta \langle \mathbf{y}_t - \mathbf{y}_t', \nabla F_{B_t}(\mathbf{w}_t') - \nabla F_{B_t}(\mathbf{v}_t') \rangle \\
&+ \eta^2 \|\nabla F_{B_t}(\mathbf{w}_t') - \nabla F_{B_t}(\mathbf{v}_t')\|^2.
\end{align*}
Since $\|\mathbf{y}_t - \mathbf{y}_t'\| \leq (1 - s\mu)\delta_t + O(sK\delta_t^2)$, we have $\|\mathbf{y}_t - \mathbf{y}_t'\|^2 \leq (1 - s\mu)^2 \delta_t^2 + O(\delta_t^3)$.
By similarity of points (for small perturbations and bounded higher-order terms),
\begin{align*}
&\langle \mathbf{y}_t - \mathbf{y}_t', \nabla F_{B_t}(\mathbf{w}_t') - \nabla F_{B_t}(\mathbf{v}_t') \rangle \\
&= \langle \mathbf{w}_t' - \mathbf{v}_t', \nabla F_{B_t}(\mathbf{w}_t') - \nabla F_{B_t}(\mathbf{v}_t') \rangle + O(\rho \delta_t^2 / G).
\end{align*}
By smoothness, $ \|\nabla F_{B_t}(\mathbf{w}_t') - \nabla F_{B_t}(\mathbf{v}_t')\| \leq L \|\mathbf{w}_t' - \mathbf{v}_t'\| \leq L (1 - s \mu) \delta_t + O(L \delta_t^2) $.
Substituting and retaining higher-order terms,
\begin{align*}
\delta_{t+1}^2 &\leq \delta_t^2 - 2\eta \frac{\mu L}{\mu + L} (1 - s \mu)^2 \delta_t^2 + \eta^2 L^2 (1 - s \mu)^2 \delta_t^2 \\
&+ O(\eta \delta_t^3 + \eta^2 \delta_t^3 + \rho \eta \delta_t^3 / G).
\end{align*}
Simplifying the leading terms,
\begin{align*}
\delta_{t+1}^2 \leq \delta_t^2 \left[ 1 - 2\eta \frac{\mu L (1 - s \mu)^2}{\mu + L} + \eta^2 L^2 (1 - s \mu)^2 \right] + O(\delta_t^3).
\end{align*}
To ensure the contraction factor is strictly less than 1, we require $ \eta < \frac{2 (1 - s \mu)^2}{\mu + L} $, yielding
\begin{align*}
\delta_{t+1} \leq \left( 1 - \eta \frac{\mu L (1 - s \mu)^2}{\mu + L} \right) \delta_t + O(\delta_t^{3/2}),
\end{align*}
where the higher-order terms are negligible for sufficiently small $ \delta_t $.
\qed

\subsection{Proof of Lemma \ref{lem:differing} (Parameter Trajectory Difference for Differing Mini-Batches)}
When mini-batches differ (probability $\frac{1}{n}$), the updates are
\begin{align*}
\mathbf{w}_{t+1} = \mathbf{y}_t - \eta \nabla F_{B_t}(\mathbf{w}_t'), \quad \mathbf{v}_{t+1} = \mathbf{y}_t' - \eta \nabla F_{B_t'}(\mathbf{v}_t').
\end{align*}
The distance is
\begin{align*}
\delta_{t+1} = \|\mathbf{y}_t - \mathbf{y}_t' - \eta (\nabla F_{B_t}(\mathbf{w}_t') - \nabla F_{B_t'}(\mathbf{v}_t'))\|. 
\end{align*}
Bounding,
\begin{align*}
\delta_{t+1} &\leq \|\mathbf{y}_t - \mathbf{y}_t' - \eta (\nabla F_{B_t}(\mathbf{w}_t') - \nabla F_{B_t}(\mathbf{v}_t'))\| \\
&+ \eta \|\nabla F_{B_t}(\mathbf{v}_t') - \nabla F_{B_t'}(\mathbf{v}_t')\|.
\end{align*}
The first term is bounded by Lemma \ref{lem:identical}'s contraction including higher-order terms:
\begin{align*}
&\|\mathbf{y}_t - \mathbf{y}_t' - \eta (\nabla F_{B_t}(\mathbf{w}_t') - \nabla F_{B_t}(\mathbf{v}_t'))\|\\
&\leq \left( 1 - \eta \frac{\mu L (1 - s \mu)^2}{\mu + L} \right) \delta_t + O(\delta_t^{3/2}).
\end{align*}
For the second term, by Lipschitz continuity and bounded gradients ($\|\nabla F_{B_t}(\cdot)\| \leq G$),
\begin{align*}
\|\nabla F_{B_t}(\mathbf{v}_t') - \nabla F_{B_t'}(\mathbf{v}_t')\|\leq 2G + O(\rho \delta_t / G + s K \delta_t^2 / G),
\end{align*}
since the mini-batches differ by at most one sample, we use the conservative upper bound $2G$. The higher-order terms arise from the perturbation normalization and Taylor remainders, Thus,
\begin{align*}
\delta_{t+1} &\leq \left( 1 - \eta \frac{\mu L (1 - s \mu)^2}{\mu + L} \right) \delta_t + 2\eta G \\
&+ O(\delta_t^{3/2} + \eta \rho \delta_t^2 / G + \eta s K \delta_t^3).
\end{align*}
\qed

\subsection{Proof of Theorem \ref{thm:gen} (Generalization Error Bound)}
The expected trajectory difference at step $ t+1 $ is given by
\begin{align*}
\mathbb{E}[\delta_{t+1}] = \left(1 - \frac{1}{n}\right) \mathbb{E}[\delta_{t+1} \mid \text{identical}] + \frac{1}{n} \mathbb{E}[\delta_{t+1} \mid \text{differing}].
\end{align*}
From Lemma \ref{lem:identical} and Lemma \ref{lem:differing}, we obtain
\begin{align*}
&\mathbb{E}[\delta_{t+1}]\leq (1 - \frac{1}{n}) [(1 - \kappa) \mathbb{E}[\delta_t] + O(\mathbb{E}[\delta_t]^{3/2})] \\
&+ \frac{1}{n} [ (1 - \kappa) \mathbb{E}[\delta_t] + 2\eta G + O(\mathbb{E}[\delta_t]^{3/2} + \eta s K \mathbb{E}[\delta_t]^3)] \\
&\leq (1 - \kappa) \mathbb{E}[\delta_t] + \frac{2\eta G}{n} + O(\mathbb{E}[\delta_t]^{3/2}).
\end{align*}
where $ \kappa = \eta \frac{\mu L (1 - s \mu)^2}{\mu + L} $ and the $ O(\cdot) $ term absorbs constants depending on $ K $, $ s $, $ \rho $, and $ G $. The additional term $O(\eta s K \mathbb{E}[\delta_t]^3)$ from the prediction remainder in differing batches is absorbed into $O(\mathbb{E}[\delta_t]^{3/2})$, as it is higher-order and negligible under small $\delta_t$ and the contraction factor $(1 - \kappa)$.
This yields the perturbed linear recurrence
\begin{align*}
\mathbb{E}[\delta_{t+1}] \leq a\mathbb{E}[\delta_t] + b + O(\mathbb{E}[\delta_t]^{3/2}),
\end{align*}
with $ a = 1 - \kappa < 1 $ (assuming $ \kappa > 0 $) and $ b = \frac{2\eta G}{n} $.
Starting from $ \delta_0 = 0 $, and assuming the initial distance is sufficiently small and the learning rate $ \eta $ is chosen small enough so that the higher-order term remains dominated by the linear contraction during the iteration process, the expected distance converges to a steady-state bound of the form
\begin{align*}
\mathbb{E}[\delta_T] &\leq \frac{b}{1 - a} + O\left( \left( \frac{b}{1 - a} \right)^{3/2} \right) \\
&= \frac{2\eta G / n}{\kappa} + O\left( \left( \frac{\eta G}{n \kappa} \right)^{3/2} \right).
\end{align*}
Substituting the expression for $ \kappa $, we obtain
\begin{align*}
\mathbb{E}[\delta_T] \leq \frac{2 G (\mu + L)}{n \mu L (1 - s \mu)^2} + O\left( \left( \frac{\eta G (\mu + L)}{n \mu L (1 - s \mu)^2} \right)^{3/2} \right).
\end{align*}
By the Lipschitz continuity assumption (Definition 3), for any test point $ z $,
\begin{align*}
&\mathbb{E} | f(\mathbf{w}_T, z) - f(\mathbf{v}_T, z) | \leq G \mathbb{E}[\delta_T] \\
&\leq \frac{2 G^2 (\mu + L)}{n \mu L (1 - s \mu)^2} + O\left( \left( \frac{\eta G^2 (\mu + L)}{n \mu L (1 - s \mu)^2} \right)^{3/2} \right).
\end{align*}
Since the bound holds uniformly over all $ z $, and uniform stability implies the generalization error bound, we conclude
\begin{align*}
\varepsilon_{\text{gen}} \leq \frac{2 G^2 (\mu + L)}{n \mu L (1 - s \mu)^2} + O\left( \left( \frac{\eta G^2 (\mu + L)}{n \mu L (1 - s \mu)^2} \right)^{3/2} \right).
\end{align*}
\qed

\subsection{Proof of Lemma \ref{lem:progress} (One-Step Progress)}
By the descent lemma from smoothness, we have
\begin{align*}
F_{B_t}(\mathbf{w}_{t+1}) &\leq F_{B_t}(\mathbf{y}_t) - \eta \nabla F_{B_t}(\mathbf{y}_t)^\top \nabla F_{B_t}(\mathbf{w}_t')
\\
&+ \frac{\eta^2 L}{2} \|\nabla F_{B_t}(\mathbf{w}_t')\|^2.
\end{align*}
We now bound $\langle \nabla F_{B_t}(\mathbf{y}_t), \nabla F_{B_t}(\mathbf{w}_t') \rangle$. Expand $\nabla F_{B_t}(\mathbf{y}_t)$ using first-order Taylor with remainder:
\begin{align*}
\nabla F_{B_t}(\mathbf{y}_t) = \nabla F_{B_t}(\mathbf{w}_t) - s \mathbf{H}_t \nabla F_{B_t}(\mathbf{w}_t) + R^{(1)}_t,
\end{align*}
where the remainder satisfies
\begin{align*}
\|R^{(1)}_t\| \leq \frac{K s^2}{2} \|\nabla F_{B_t}(\mathbf{w}_t)\|^2,
\end{align*}
under the Hessian Lipschitz assumption (constant $K$). Then, the perturbed gradient is
\begin{align*}
\nabla F_{B_t}(\mathbf{w}_t') = \nabla F_{B_t}(\mathbf{y}_t) + \rho \frac{\nabla F_{B_t}(\mathbf{w}_t)}{\|\nabla F_{B_t}(\mathbf{w}_t)\|}.  
\end{align*}
Let $\mathbf{u}_t = \frac{\nabla F_{B_t}(\mathbf{y}_t)}{\|\nabla F_{B_t}(\mathbf{y}_t)\|}$. Using the fact that the map $\mathbf{g} \mapsto \mathbf{g}/\|\mathbf{g}\|$ is Lipschitz on the sphere with constant bounded by $1/G$ (assuming $\|\nabla F_{B_t}(\mathbf{y}_t)\| \geq G > 0$), the perturbation term introduces an additional error $O(\rho \|R^{(1)}_t\| / G) = O(\rho s^2 \|\nabla F_{B_t}(\mathbf{w}_t)\|^2 / G)$.
Substituting, we obtain
\begin{align*}
\nabla F_{B_t}(\mathbf{w}_t') &= (I - s \mathbf{H}_t) \nabla F_{B_t}(\mathbf{w}_t) + R^{(1)}_t + \rho \mathbf{u}_t \\
&+ O(\rho s^2 \|\nabla F_{B_t}(\mathbf{w}_t)\|^2 / G).
\end{align*}
The inner product becomes
\begin{align*}
&\langle \nabla F_{B_t}(\mathbf{y}_t), \nabla F_{B_t}(\mathbf{w}_t') \rangle = \nabla F_{B_t}(\mathbf{w}_t)^\top (I - s \mathbf{H}_t) \nabla F_{B_t}(\mathbf{w}_t) \\
&+ \nabla F_{B_t}(\mathbf{w}_t)^\top R^{(1)}_t + \rho \nabla F_{B_t}(\mathbf{w}_t)^\top \mathbf{u}_t \\
&+ O(\rho s^2 \|\nabla F_{B_t}(\mathbf{w}_t)\|^3 / G).
\end{align*}
Using the Taylor expansion 
\begin{align*}
\nabla F_{B_t}(\mathbf{y}_t) = (I - s \mathbf{H}_t) \nabla F_{B_t}(\mathbf{w}_t) + R^{(1)}_t, 
\end{align*}
the above inner product satisfies
\begin{align*}
\nabla F_{B_t}(\mathbf{w}_t)^\top \mathbf{H}_t \nabla F_{B_t}(\mathbf{w}_t) &\geq \mu \|\nabla F_{B_t}(\mathbf{w}_t)\|^2,\\
\nabla F_{B_t}(\mathbf{w}_t)^\top (I - s \mathbf{H}_t) \nabla F_{B_t}(\mathbf{w}_t)&\geq (1 - s \mu) \|\nabla F_{B_t}(\mathbf{w}_t)\|^2.
\end{align*}
The remainder terms are bounded as
\begin{align*}
\|\nabla F_{B_t}(\mathbf{w}_t)^\top R^{(1)}_t\| &\leq \|\nabla F_{B_t}(\mathbf{w}_t)\| \cdot \|R^{(1)}_t\| \\
&\leq \frac{K s^2}{2} \|\nabla F_{B_t}(\mathbf{w}_t)\|^3,
\end{align*}
and similarly for the other higher-order terms. Assuming bounded gradients $\|\nabla F_{B_t}(\mathbf{w}_t)\| \leq G$, the overall inner product is
\begin{align*}
\langle \nabla F_{B_t}(\mathbf{y}_t), \nabla F_{B_t}(\mathbf{w}_t') \rangle &\geq (1 - s \mu) \|\nabla F_{B_t}(\mathbf{w}_t)\|^2\\
&- O(s^2 G^3 + \rho s^2 G^2).
\end{align*}
From strong convexity,
\begin{align*}
\|\nabla F_{B_t}(\mathbf{w}_t)\|^2 &\geq 2\mu (F_{B_t}(\mathbf{w}_t) - F_{B_t}(\mathbf{w}^*)),\\
\langle \nabla F_{B_t}(\mathbf{y}_t), \nabla F_{B_t}(\mathbf{w}_t') \rangle &\geq 2\mu (1 - s \mu) (F_{B_t}(\mathbf{w}_t) - F_{B_t}(\mathbf{w}^*)) \\
&- O(s^2 G^3).
\end{align*}
Substituting back into the descent inequality, and bounding 
\begin{align*}
\|\nabla F_{B_t}(\mathbf{w}_t')\|^2 \leq (G + O(s G^2 + \rho G))^2 \approx G^2 + O(s G^3 + \rho G^2),
\end{align*}
hence
\begin{align*}
&F_{B_t}(\mathbf{w}_{t+1}) - F_{B_t}(\mathbf{w}^*)\\
&\leq (1 - \eta \cdot 2\mu (1 - s \mu)) (F_{B_t}(\mathbf{w}_t) - F_{B_t}(\mathbf{w}^*)) + \frac{\eta^2 L G^2}{2} \\
&+ O(\eta s^2 G^3 + \eta \rho G^2).
\end{align*}
Let \(\alpha = 2\mu (1 - s \mu)\). Then
\begin{align*}
F_{B_t}(\mathbf{w}_{t+1}) - F_{B_t}(\mathbf{w}^*) &\leq (1 - \eta \alpha) (F_{B_t}(\mathbf{w}_t) - F_{B_t}(\mathbf{w}^*)) \\
&+ \frac{\eta^2 L G^2}{2} + O(\eta s^2 G^3 + \eta \rho G^2).
\end{align*}
\qed

\subsection{Proof of Theorem \ref{thm:opt} (Optimization Error Bound)}
From the updated Lemma \ref{lem:progress}, taking expectations over mini-batches, we have
\begin{align*}
&\mathbb{E}[F_{B_t}(\mathbf{w}_{t+1}) - F_{B_t}(\mathbf{w}^*)] \\
&\leq (1 - \eta \cdot 2\mu (1 - s \mu)) \mathbb{E}[F_{B_t}(\mathbf{w}_t) - F_{B_t}(\mathbf{w}^*)] + \frac{\eta^2 L G^2}{2} \\
&+ O(\eta s^2 G^3 + \eta \rho G^2).
\end{align*}
This yields the recurrence
\begin{align*}
e_{t+1} \leq (1 - \eta \alpha) e_t + c + O(\eta s^2 G^3 + \eta \rho G^2),
\end{align*}
where $e_t = \mathbb{E}[F_{B_t}(\mathbf{w}_t) - F_{B_t}(\mathbf{w}^*)]$, $\alpha = 2\mu (1 - s \mu)$, and $c = \frac{\eta^2 L G^2}{2}$.
Assuming the initial suboptimality $e_0$ is bounded and the higher-order terms are dominated by the linear contraction (via sufficiently small $s$, $\rho$, and appropriate $\eta$), iterating the recurrence gives, for large $T$,
\begin{align*}
e_T &\leq \frac{c}{\eta \alpha} + O\left( \eta s^2 G^3 + \eta \rho G^2 \right) \\
&= \frac{\eta L G^2}{4 \mu (1 - s \mu)} + O\left( \eta s^2 G^3 + \eta \rho G^2 \right).
\end{align*}
Thus,
\begin{align*}
\varepsilon_{\text{opt}} \leq \frac{\eta L G^2}{4 \mu (1 - s \mu)} + O\left( \eta s^2 G^3 + \eta \rho G^2 \right).
\end{align*}
\qed

\subsection{Proof of Theorem \ref{thm:exc} (Expected Excess Risk Bound)}
By the standard decomposition of excess risk into generalization error and optimization error (neglecting the approximation error for large $n$),
\begin{align*}
\varepsilon_{\text{exc}} \leq \varepsilon_{\text{gen}} + \varepsilon_{\text{opt}}.
\end{align*}
Substituting the bounds from the updated Theorems \ref{thm:gen} and \ref{thm:opt},
\begin{align*}
\varepsilon_{\text{exc}} &\leq \left[ \frac{2 G^2 (\mu + L)}{n \mu L (1 - s \mu)^2} + O\left( \left( \frac{\eta G^2 (\mu + L)}{n \mu L (1 - s \mu)^2} \right)^{3/2} \right) \right] \\
&+ \left[ \frac{\eta L G^2}{4 \mu (1 - s \mu)} + O(\eta s^2 G^3 + \eta \rho G^2) \right].
\end{align*}
Under the learning rate constraint $\eta \leq \frac{2 (1 - s \mu)^2}{\mu + L}$, the higher-order terms remain controlled and do not dominate the leading terms for sufficiently small $s$, $\rho$, and large $n$. The leading generalization term benefits from the prediction step factor $(1 - s \mu)^2$, while the optimization term is tightened by $(1 - s \mu)$ in the denominator. The perturbation $\rho$ contributes only to higher-order errors, which are negligible under appropriate hyperparameter tuning.
\qed

\subsection{Proof of Theorem \ref{thm:Nonconvex exc} (Nonconvex Expected Excess Risk Bound)}
We prove the theorem by decomposing the expected excess risk $\mathbb{E}[F(\bar{\mathbf{w}}_T) - F(\mathbf{w}^*)]$ into optimization error, sharpness control term, and generalization gap, where $\bar{\mathbf{w}}_T = \frac{1}{T} \sum_{t=1}^T \mathbf{w}_t$.

By $L$-smoothness of $F$, the descent lemma applied to $\mathbf{y}_t$ yields
\begin{align*}
F(\mathbf{w}_{t+1}) \leq F(\mathbf{y}_t) - \eta \|\nabla F(\mathbf{w}_t')\|^2 + \frac{\eta^2 L}{2} \mathbb{E}[\|\nabla f(\mathbf{w}_t'; z_t)\|^2].
\end{align*}
The prediction step satisfies
\begin{align*}
F(\mathbf{y}_t) &\leq F(\mathbf{w}_{t-1}) - s \|\nabla F(\mathbf{w}_{t-1})\|^2 \\
&+ \frac{s^2 L}{2} \mathbb{E}[\|\nabla f(\mathbf{w}_{t-1}; z_{t-1})\|^2].
\end{align*}

Since the prediction phase performs an actual parameter update $\mathbf{w}_t \leftarrow \mathbf{y}_t$, for notational convenience in the telescoping sum we identify $\mathbf{w}_t \equiv \mathbf{y}_t$. Combining both phases with expectation, we obtain
\begin{align*}
\mathbb{E}[F(\mathbf{w}_{t+1})]
&\leq \mathbb{E}[F(\mathbf{w}_t)]
- \eta (1 - sL - \rho L) \mathbb{E}[\|\nabla F(\mathbf{w}_t)\|^2] \\
&\quad + \eta^2 L \sigma^2 + s^2 L \sigma^2 + O(\rho^2 L).
\end{align*}
By choosing the learning rate $\eta$ sufficiently small so that $1 - sL - \rho L > 0$, the coefficient remains positive. Thus we still obtain sufficient descent.
\begin{align*}
\frac{1}{T} \sum_{t=1}^T \mathbb{E}[\|\nabla F(\mathbf{w}_t)\|^2] 
&\leq \frac{F(\mathbf{w}_1) - F^*}{\eta T (1 - sL - \rho L)} 
+ \frac{\eta L \sigma^2 + s^2 L \sigma^2}{1 - sL - \rho L} \\
&+ O\left(\frac{\rho^2 L}{\eta}\right).
\end{align*}
Applying Jensen's inequality and $L$-smoothness to the averaged parameters $\bar{\mathbf{w}}_T$ gives
\begin{align*}
\mathbb{E}[F(\bar{\mathbf{w}}_T)] 
&\leq F(\mathbf{w}_1) 
- \frac{\eta (1 - s L - \rho L)}{2} \sum_{t=1}^T \mathbb{E}[\|\nabla F(\mathbf{w}_t)\|^2] \\
&+ \frac{\eta^2 L T \sigma^2 + s^2 L T \sigma^2}{2}.
\end{align*}
This establishes the optimization error term
\begin{align*}
\mathcal{O}\left( \frac{F(\mathbf{w}_1) - F^*}{\eta T (1 - s L - \rho L)} + \eta L \sigma^2 + s^2 L \sigma^2 + \rho^2 L \right).
\end{align*}
For the sharpness control term, the perturbation $\boldsymbol{\epsilon}_t$ explicitly minimizes neighborhood sharpness. Under $L$-smoothness, the second-order Taylor expansion bounds
\begin{align*}
\max_{\|\epsilon\| \leq \rho} F(\mathbf{w}_t + \epsilon) 
&\approx F(\mathbf{w}_t) + \rho \|\nabla F(\mathbf{w}_t)\| \\
&+ \frac{\rho^2}{2} \lambda_{\max}(\nabla^2 F(\mathbf{w}_t)),
\end{align*}
yielding a controlled $O(\rho^2 L)$ term (tighter for large $\rho$ due to correction stability from $\mathbf{y}_t$).

For the generalization gap, we employ the algorithmic stability framework of Hardt et al. \cite{Hardt2016}. Let $(\mathbf{w}_t^{(1)})$ and $(\mathbf{w}_t^{(2)})$ be two parallel executions of EISAM on identical mini-batches $z_t$ but with independent stochastic gradient noise. Define $\delta_t := \mathbb{E}[\|\mathbf{w}_t^{(1)} - \mathbf{w}_t^{(2)}\|_2^2]$. The difference evolves as
\begin{equation*}
\mathbf{w}_{t+1}^{(1)} - \mathbf{w}_{t+1}^{(2)} = (\mathbf{y}_t^{(1)} - \mathbf{y}_t^{(2)}) - \eta (\nabla f(\mathbf{w}_t'^{(1)}; z_t) - \nabla f(\mathbf{w}_t'^{(2)}; z_t)).
\end{equation*}
Expanding the squared norm and expectation, the prediction contribution satisfies
\begin{align*}
\mathbb{E}[\|\mathbf{y}_t^{(1)} - \mathbf{y}_t^{(2)}\|_2^2] 
\leq (1 + s^2 L^2)\delta_t + O(s^2 \sigma^2)
\end{align*}
by $L$-smoothness. On the perturbed points $\mathbf{w}_t'^{(i)}$, $L$-smoothness together with a local Polyak--Łojasiewicz inequality implies a contraction term of order $-2\eta\mu \mathbb{E}[\|\mathbf{y}_t^{(1)} - \mathbf{y}_t^{(2)}\|_2^2]$. The quadratic variance term is bounded by $2\eta^2\sigma^2$. Collecting terms yields the contraction
\begin{align*}
\delta_{t+1} \leq (1 - c\eta\mu)\delta_t + O(\eta^2 \sigma^2 + s^2 \sigma^2), \quad c \in (0,1).
\end{align*}
Solving the recurrence gives
\begin{align*}
\delta_T = O\left(\frac{\sigma^2}{\mu}\right) + O\left(\frac{\sigma^2 (1 + s^2 L^2)}{\eta \mu}\right).
\end{align*}
By the standard conversion from mean-square stability to uniform stability \cite{Hardt2016}, the expected generalization gap satisfies
\begin{align*}
\mathbb{E}[F_{\mathcal{D}}(\mathbf{w}_T) - F_S(\mathbf{w}_T)] 
= O(\sigma \sqrt{d \log T / n}).
\end{align*}
Combining the optimization error, sharpness control term, and generalization gap yields the claimed nonconvex expected excess risk bound
\begin{align*}
\varepsilon_{\rm exc} &\leq \mathcal{O}\left( \frac{F(\mathbf{w}_1) - F^*}{\eta T (1 - s L - \rho L)} \right)\\
&+ \mathcal{O}\left(\eta L \sigma^2+s^2 L \sigma^2 + \rho^2 L + \sigma \sqrt{\frac{d \log T}{n}} \right).
\end{align*}
The dominant optimization term is sublinear in $T$, while the generalization term decays as $\mathcal{O}(1/\sqrt{n})$.
\qed

\begin{figure}[htbp]
    \centering
    \includegraphics[width=0.47
    \textwidth, angle=0, trim=0.1cm 0.1cm 0.1cm 0.1cm, clip]{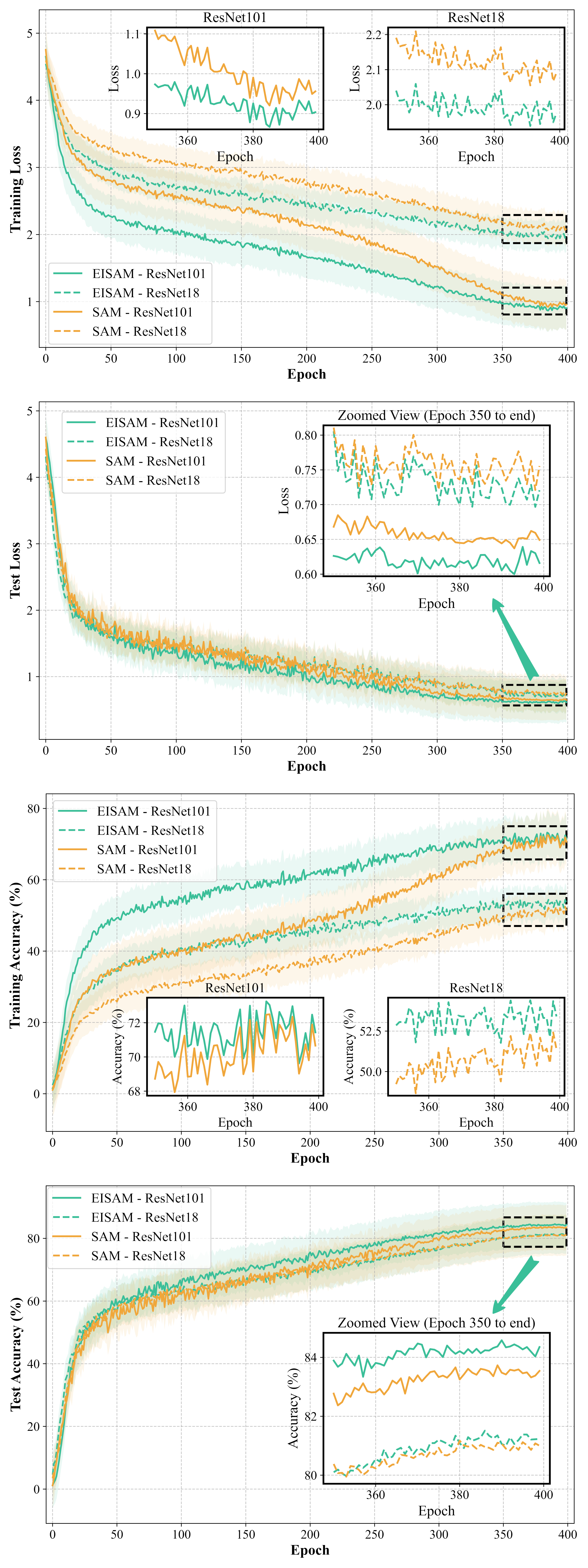}
    \caption{{Training loss , training accuracy, test loss as well as test accuracy of ResNet18 and ResNet101 on CIFAR-100 with CutMix Augmentation(Expanded to 400 epochs)}.}
    \label{long}
\end{figure}

\section{Extended Training Curves for Long-Term Optimization Behavior of SAM and EISAM}
\label{appc}

To further investigate the long-term optimization dynamics of SAM and EISAM, we extend the experiments presented in Fig. \ref{F5} of the main paper to 400 epochs, as illustrated in the accompanying Fig. \ref{long}. These curves compare EISAM and SAM across training loss, test loss, training accuracy, and test accuracy.

The results demonstrate that EISAM consistently outperforms SAM even over prolonged training. In terms of training loss, EISAM achieves faster convergence and lower final values. For test loss, EISAM maintains a clear advantage, reaching substantially lower values and showing reduced variance in the terminal phase. This indicates superior generalization and robustness to overfitting during extended optimization.
Regarding training accuracy, EISAM rapidly surpasses SAM and sustains higher performance. Most notably, the test accuracy curves reveal EISAM’s superior generalization: it achieves higher final accuracy, continues to improve steadily in the later epochs. In contrast, SAM exhibits larger oscillations and plateaus earlier, suggesting it struggles to escape suboptimal regions in prolonged training.

Overall, these extended experiments confirm that EISAM not only converges more efficiently but also yields better final performance, enhanced stability, and stronger generalization compared to SAM, even when trained for significantly longer periods. This underscores EISAM’s robustness in long-horizon deep learning optimization scenarios. Additionally, EISAM demonstrates superior convergence speed compared to SAM, thereby enabling faster hyperparameter tuning and reducing overall training time.

\section{Comprehensive Comparison of All Experiments}
\label{all}
\begin{figure*}[htbp]
    \centering
    \includegraphics[width=1
    \textwidth, angle=0, trim=0.1cm 0.1cm 0.1cm 0.1cm, clip]{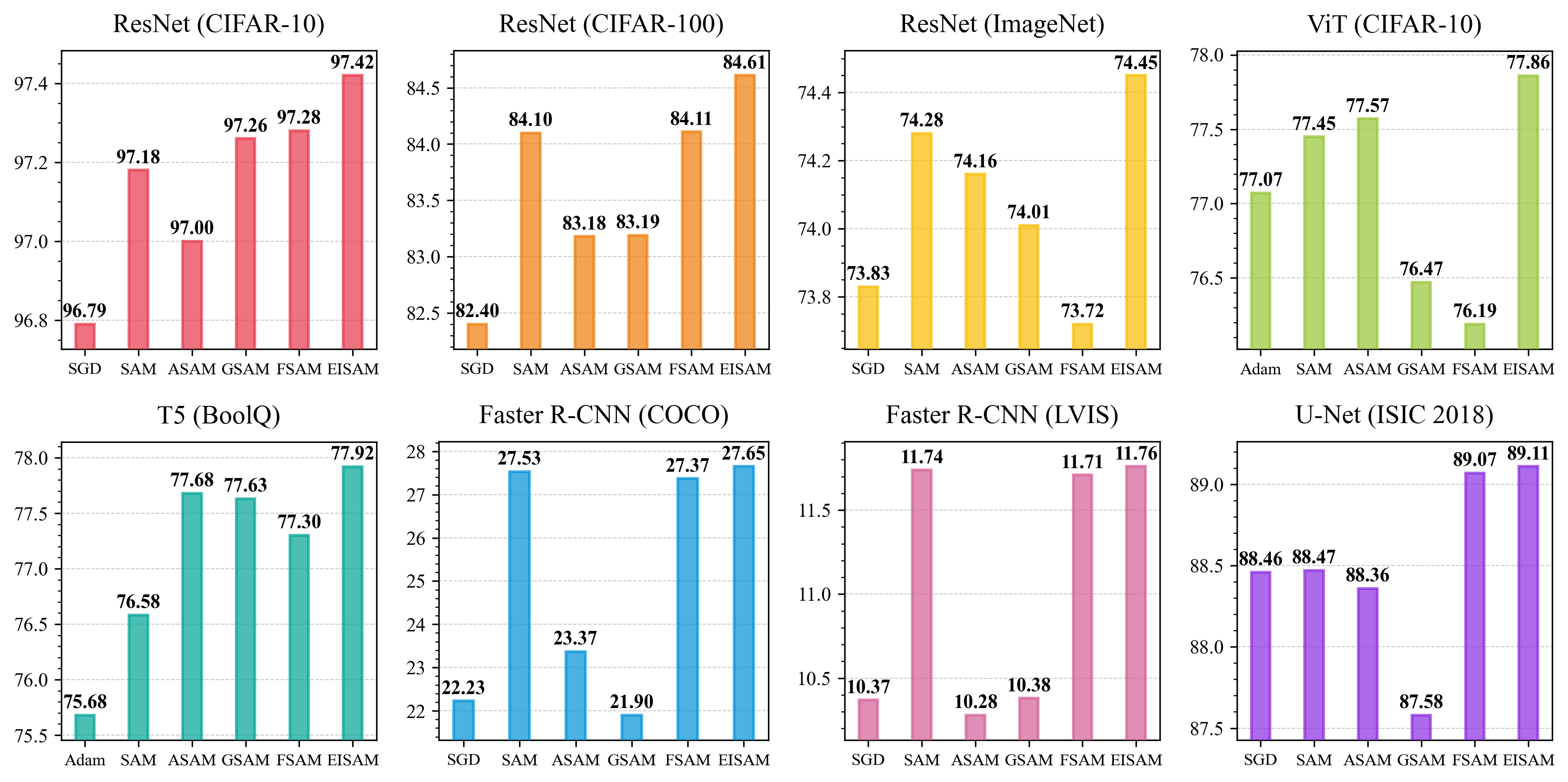}
    \caption{Comprehensive comparison of all experiments, the vertical axis shows the mean performance metric per experimental category.}
    \label{huiz}
\end{figure*}

This section provides a detailed aggregation of experimental results across all tasks, datasets, and optimizers discussed in the main paper. Fig. \ref{huiz} summarizes key performance metrics, allowing for a direct visual comparison of generalization ability among the evaluated methods. All reported results are the averages of the performance metrics from the experimental tables in Section \ref{sec5}. These aggregated numbers further confirm the consistent advantages of EISAM across diverse settings and architectures.

\section{Computational Cost Comparison of All Optimizers}
\label{cost} 

To more comprehensively evaluate the performance of EISAM, we further recorded and analyzed its computational overhead along with that of the baseline optimizers (SGD, SAM, ASAM, GSAM, and FSAM) in object detection experiments. Specifically, we employed a pretrained Faster R-CNN model with a ResNet-50-FPN backbone on the LVIS V1.0 dataset, setting the batch size to 6 (without gradient accumulation) and conducting a complete training run to record various metrics for each optimizer. The hardware environment consisted of an AMD Ryzen 9 7950X processor paired with an RTX 5090 GPU. The experimental results are presented in Table \ref{tab:cost}, where time-related metrics are reported in seconds, memory-related metrics in MB, and FLOPs in Tera (T).

\begin{table*}[!t]
\caption{Computational cost comparison}
\label{tab:cost}
\centering
\normalsize
\begin{tabular}{lcccccc}
\toprule
Metric & SGD & SAM & ASAM & GSAM & FSAM & EISAM(Ours) \\
\midrule
Weight Update Time & 0.0020 & 0.2835 & 0.2861 & 0.2915 & 0.2935 & 0.2957 \\
Per-batch Time & 0.2673 & 0.5519 & 0.5543 & 0.5593 & 0.5645 & 0.5615 \\
Total Time & 4515.1982 & 9286.6064 & 9326.1706 & 9410.3044 & 9497.2772 & 9449.1931 \\
Weight Update Mem. & 360.3926 & 16726.1880 & 16726.1880 & 16545.9917 & 16726.1880 & 16726.1880 \\
Optimizer State Mem. & 238.4209 & 238.4209 & 238.4209 & 598.8135 & 418.6172 & 598.8135 \\
Peak Total Mem. & 17696.0 & 18040.0 & 18420.0 & 18274.0 & 18702.0 & 18540.0 \\
Per-batch FLOPs & 8.9328 & 17.8656 & 17.8656 & 17.8656 & 17.8656 & 17.8656 \\
\bottomrule
\end{tabular}
\end{table*}

\begin{table*}[htbp]
\centering
\caption{Hyperparameters for Different Models and Datasets}
\label{tab:tuning}
\normalsize
\begin{tabular}{lccccc}
\toprule
Model (Dataset) & Learning rate & Weight decay & Batch size & EISAM-$\rho$ & EISAM-s \\
\midrule
ResNet18 (CIFAR-100) & 0.05 & 1e-3 & 128 & 0.1 & 5e-3 \\
ResNet50 (CIFAR-100) & 0.05 & 1e-3 & 128 & 0.05 & 1e-3 \\
ResNet101 (CIFAR-100) & 0.05 & 1e-3 & 128 & 0.1 & 1e-3 \\
WideResNet (CIFAR-100) & 0.05 & 1e-3 & 128 & 0.1 & 1e-3 \\
PyramidNet (CIFAR-100) & 0.05 & 1e-3 & 128 & 0.1 & 1e-3 \\
\midrule
ResNet18 (ImageNet) & 0.1 & 1e-4 & 256 & 0.01 & 5e-3 \\
ResNet50 (ImageNet) & 0.1 & 1e-4 & 256 & 0.01 & 5e-3 \\
\midrule
ViT-S-16-Cutmix (CIFAR-10) & 1e-4 & 0 & 32 & 5e-3 & 1e-4 \\
\midrule
T5 (BOOLQ) & 5e-4 & 5e-4 & 64 & 0.05 & 1e-4 \\
\midrule
F-RCNN (LVIS) & 1e-3 & 1e-4 & 6 & 0.05 & 5e-4 \\
F-RCNN (COCO) & 1e-3 & 1e-4 & 5 & 0.05 & 5e-4 \\
\midrule
U-Net (ISIC2018) & 5e-4 & 5e-4 & 512 & 1e-3 & 1e-4 \\
\bottomrule
\end{tabular}
\end{table*}

\begin{itemize}
    \item Weight Update Time: This metric quantifies the time consumed by each optimizer for parameter updates according to its own optimization mechanism after the first forward and backward pass (i.e., the execution time of \texttt{optimizer.step()} in the PyTorch framework). SAM-family optimizers require an additional forward-backward pass and therefore incur substantially higher cost in this phase compared with SGD; this component best highlights the incremental overhead introduced by each variant’s specific design relative to vanilla SAM.
    \item Per-batch Time: To faithfully capture pure wall-clock training time, we disabled the evaluation process during training. This metric nevertheless includes minor overhead from operations such as CUDA synchronization (\texttt{torch.cuda.synchronize()}) required for accurate timing and memory measurement.
    \item Total Time: Although the evaluation phase was omitted, the total time still accounts for small additional costs from CUDA synchronization, data transfer, and disk I/O. To minimize extraneous overhead, data were loaded directly into pinned memory and the process was kept alive throughout execution.
    \item Weight Update Memory: This records the peak memory usage during the parameter-update phase (before and after \texttt{optimizer.step()}). SAM-family optimizers occupy more memory here; however, because this phase executes sequentially after the initial forward-backward pass, the additional memory does not accumulate linearly with the memory reserved for the first pass.
    \item Optimizer State Memory: This reflects the static memory footprint occupied by each optimizer’s internal state. It was computed after training completion (with data, gradients, and activations cleared) by subtracting the model-parameter memory from the total reserved memory.
    \item Peak Total Memory: This is the maximum memory reserved by CUDA during the entire training process. We recorded \texttt{torch.cuda.max\_memory\_reserved()} before and after each stage of every batch iteration to obtain the most accurate peak values.
    \item Per-batch FLOPs: Measured by wrapping the full per-batch iteration with PyTorch’s built-in \texttt{FlopCounterMode} module, accumulating operations across all steps, and dividing by the total number of iterations.
\end{itemize}

The wall-clock time and peak memory consumption of EISAM are on par with those of ASAM, GSAM, and FSAM. Relative to SAM, EISAM increases per-batch wall-clock time by only 4.4\% and total training time by only 1.7\%, while the peak memory footprint rises by merely 2.7\%. Notably, the FLOPs per batch remain identical (17.8656 T) across all SAM variants, confirming that EISAM introduces no additional computational complexity in the core gradient-evaluation steps.

\section{Hyperparameter Tuning Guidelines for EISAM}
\label{hyper}

EISAM introduces two primary hyperparameters beyond those of the base optimizer: the sharpness parameter $  \rho  $ and the prediction step size $  s  $. Proper tuning of these parameters is essential for achieving optimal generalization and convergence. Below, we provide practical guidelines based on our experiments across benchmark datasets and architectures.

Sharpness Parameter $  \rho  $ controls the magnitude of the sharpness-aware perturbation, directly influencing the optimizer's bias toward flatter minima.
Recommended range: 0.001 to 0.2 (default: 0.05, as in standard SAM).

Prediction Step Size $  s  $ determines the length of the initial prediction step, which probes the local geometry of the loss landscape. Recommended range: 1e-5 to 0.01 (default: 0.001). Smaller $  s  $ yields more accurate geometry probing. Larger $  s  $can accelerate convergence but risk overshooting in ill-conditioned landscapes. For most tasks, directly adopting the default values and adding $s$ cosine scheduling is sufficient (as embodied in the article's code). Reduce $  s  $ if training becomes unstable.

The hyperparameters of the base optimizer, such as learning rate, batch size, and weight decay, can be tuned along with the base optimizer. The recommended hyperparameters for the key experimental settings are summarized in Table \ref{tab:tuning}.



%



\ifCLASSOPTIONcaptionsoff
  \newpage
\fi

\end{document}